\documentclass[twoside,letterpaper]{article}
\usepackage[T1]{fontenc}
\usepackage{aistats2027,times}
\makeatletter
\renewcommand{\statePaper}{\acceptedPaper}
\renewcommand{\Notice@String}{Preprint. Under review.}
\makeatother
\runningtitle{Preprint. Under review.}
\runningauthor{Preprint. Under review.}
\usepackage[round]{natbib}
\renewcommand{\bibname}{References}
\renewcommand{\bibsection}{\subsubsection*{\bibname}}
\usepackage{amsmath,amssymb}
\usepackage{booktabs}
\usepackage{multirow}
\usepackage{graphicx}
\usepackage{microtype}
\usepackage[hidelinks]{hyperref}
\usepackage{url}
\SetProtrusion{encoding=T1,family=pcr}{}

\newcommand{\manuscripttitle}{Post-Grokking Collapse at the Representation--Readout Interface\\ in Muon-Trained Transformers}

\begin{document}

\twocolumn[
\aistatstitle{\manuscripttitle}
\vspace{-24.052pt}
\aistatsauthor{Ali Janati\textsuperscript{1,2}\quad Kaoutar El Maghraoui\textsuperscript{2}\quad Anass Belfatmi\textsuperscript{3}}
\aistatsaddress{\small\textsuperscript{1}Data Science Institute, Columbia University, New York, NY, USA\\
\small\textsuperscript{2}Department of Computer Science, Columbia University, New York, NY, USA\\
\small\textsuperscript{3}CentraleSup\'elec, Gif-sur-Yvette, France}
]

\begin{abstract}
Muon-trained modular-arithmetic transformers can lose accuracy while retaining linearly decodable task information. Adjacent swaps localize five captured unnormalized failures to AdamW readout updates. Multiplying the actual readout displacement by the large feature mean produces a class-dependent logit offset shared across inputs that nearly reproduces each failure. Training-only decoders recover $98.20$--$100\%$ held-out accuracy. Correcting cross-entropy derivative errors stabilizes five matched branches through step $100{,}000$; four prospective accurate-CE RMS runs fail through embedding updates.
\end{abstract}

\section{Introduction}
\label{sec:intro}

\begin{figure*}[!t]
\centering
\includegraphics[width=6.5in]{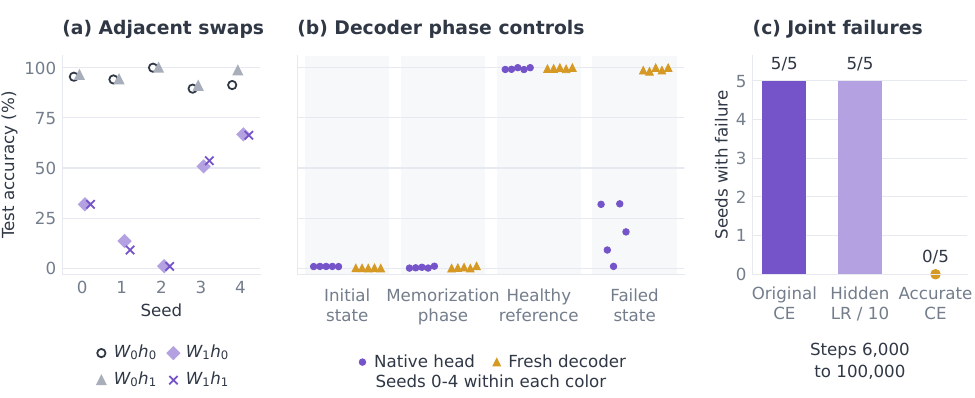}
\caption{Follow-ups on five selected unnormalized trajectories; panels reuse seeds. \textbf{(a)} Adjacent swaps: $0/1$ denote states before/after the captured update. The old head retains accuracy on new features; the new head lowers accuracy on old features. \textbf{(b)} Native-head and training-only decoder test accuracy at initialization, sustained memorization, a healthy reference, and a failed checkpoint; five seeds are shown at each state. Checkpoint identities and fitting details are in Appendix~\ref{app:decoder-controls}. Failed snapshots for seeds $3$ and $4$ occur after the adjacent transitions in (a). \textbf{(c)} Joint train/test failures below $90\%$: original CE $5/5$, hidden LR$/10$ $5/5$, accurate CE $0/5$. Continuations share step-$6{,}000$ sources and end at $100{,}000$. Binary incidence compares mixed original monitoring (failures captured before $30{,}000$) with complete joint-event detection in both continuations (Appendix~\ref{app:followup}).}
\label{fig:followup}
\end{figure*}

Generalization acquired during grokking can later collapse. In five captured failures of unnormalized modular-arithmetic transformers, swaps across a single optimizer update localize the acute damage to the readout: the preceding head remains accurate on the updated features, while the new head lowers accuracy on the preceding features. Fresh linear decoders fitted on training examples recover $98.20$--$100\%$ held-out accuracy at selected failed checkpoints. The mean-dependent component of the actual readout displacement nearly reproduces each failure (Figure~\ref{fig:followup}; \S\ref{sec:arithmetic}).

Near zero training loss, stock float32 cross-entropy can violate the logit derivative's zero class sum, adding a class-common error component. Saved-state Adam diagnostics measure its induced class-dependent update difference. Corrections begun earlier in training prevent these failures; accurate-gradient substitution only at the failing update leaves all five intact. The full upstream connection remains unresolved (\S\ref{sec:arithmetic}).

\citet{liu2026glitching} already connect inaccurate cross-entropy derivatives, feature/classifier-mean growth, amplified Adam updates, and numerical corrections. \citet{chou2026twospeeds} use a representation--readout decomposition and linear probes to diagnose readout miscalibration during grokking. We add a joint account of individual post-grokking failures: adjacent-update localization, an intervention on the mean-dependent part of the measured readout displacement, and a test of the task information retained at failed states. These interventions identify the damaging update; the upstream contributions of feature growth and optimizer history remain coupled.

We use Muon \citep{jordan2024muon} with the routing used in large-scale training \citep{liu2025muon}: Muon trains the hidden matrices, while AdamW trains the token and position embeddings and the readout. Our speed and stability comparisons concern the stated configurations, sweeps, and horizons. Matched corrected-cross-entropy continuations remain free of joint train/test failure through step $100{,}000$ in all five unnormalized seeds; a tenfold post-grokking reduction in the hidden learning rate leaves failures in all five. With gain-free RMS, accurate-CE failures replicate in four prospective seeds. Swaps localize each captured transition to embeddings, while failed-state decoder accuracy varies substantially, establishing an architectural boundary for this account.

Fourier measurements support this analysis with a separate qualification. Retaining a full task-aligned family averages representations over inputs with the same answer, supplying enough task structure for a lookup-table memorizer to achieve perfect filtered accuracy. Raw-feature decoders and adjacent swaps assess surviving information independently of this projection. Spectral similarity during a captured collapse is discussed alongside the projection analyses in \S\ref{sec:outvoted}.

\textbf{Contributions.}
\begin{itemize}
\item \textbf{Localized failures with decodable information.} At five captured unnormalized transitions, adjacent swaps isolate the readout update. Its mean-dependent displacement nearly reproduces the loss of accuracy, while fresh training-only decoders recover high \mbox{held-out accuracy} (\S\ref{sec:localize}).
\item \textbf{Stabilization and its scope.} Matched continuations compare numerical correction and hidden-learning-rate reduction through $100{,}000$ steps. Four prospective RMS runs establish limits to numerical prevention, readout localization, and retained decodability (\S\ref{sec:arithmetic}).
\item \textbf{Interpretation of Fourier interventions.} An orbit-average identity and memorizer controls explain the task structure supplied by filtering and calibrate its use as evidence for a learned algorithm (\S\ref{sec:family}).
\end{itemize}

\section{Setup}
\label{sec:setup}

\textbf{Task.} We train on $(a+b)\bmod p$, and on $(a-b)\bmod p$ where stated, using token sequences $[a,b,=]$ and $p$ output classes. Defaults are $p=113$ and a fixed random $30\%$ training split of the $p^2$ operand grid. Training is full batch with cross-entropy. Original sweeps evaluate every $100$ steps; diagnostic branches use finer grids where specified. The arithmetic follow-ups add the monitoring in \S\ref{sec:arithmetic}.

\textbf{Architecture.} The model is a decoder-only transformer with one block unless otherwise specified, $d_{\text{model}}=128$, four heads, MLP width $512$, ReLU, sequence length $3$, every projection bias-free, and no normalization layers. An invertible change of residual basis can be absorbed by the matrices that write to and read from it. This is a symmetry of the parameterization; optimizer updates need not preserve it. The gain-free RMS control restricts this symmetry to $O(d)$. Depth is varied by appending blocks under nested initialization.

\begin{table*}[!t]
\centering
\small
\caption{Parameter groups and routing. The AdamW baseline places all parameters under one AdamW instance. Selected settings: Muon hidden $\mathrm{lr}=0.03$, $\mathrm{wd}=0.1$, $\beta=0.95$, five Newton--Schulz iterations; auxiliary $\mathrm{lr}=10^{-3}$, $\mathrm{wd}=1.0$; readout $\mathrm{lr}=2.5\times10^{-4}$, $\mathrm{wd}=1.0$; AdamW baseline $\mathrm{lr}=10^{-3}$, $\mathrm{wd}=3.0$. Selection protocol and full sweeps in Appendix~\ref{app:sweeps}.}
\label{tab:routing}
\begin{tabular}{lll}
\toprule
Group & Parameters & Optimizer \\
\midrule
Hidden & QKV, attention output, MLP input, MLP output, per block & Muon \\
Auxiliary & Token embedding, position embedding & AdamW \\
Readout & Unembedding & AdamW \\
\bottomrule
\end{tabular}
\end{table*}

\textbf{Routing and definitions.} Parameters split into the three groups of Table~\ref{tab:routing}; Muon receives the hidden group. Original grokking speeds use the first of five consecutive evaluations at or above $95\%$ test accuracy. \emph{Strictly stable} means no subsequent evaluation below $95\%$; we also report sub-threshold counts and minima. The depth-$4$ freeze and fresh follow-ups verify shared source states. Historical branch provenance is reported separately (Appendix~\ref{app:control-provenance}). CPU seed replications have identical logged metrics before the freeze (Appendix~\ref{app:setup}). Follow-up confirmation uses six checks (Appendix~\ref{app:followup-protocol}).

\section{Earlier grokking in the tested Muon configurations}
\label{sec:speed}

The configuration-selection criteria differ. AdamW uses the fastest strictly stable configuration in its single-seed sweep; every tested Muon configuration is unstable, so Muon uses the fastest of those configurations (Appendix~\ref{app:sweeps}). Across five seeds at these selected settings, median grokking steps are $8{,}300$ for AdamW and $5{,}400$ for Muon, a $1.54\times$ ratio. The fastest successful single-seed sweep entries reach the criterion at $6{,}300$ and $5{,}400$ steps, respectively, a $1.17\times$ ratio. These summarize different comparisons.

All nine entries in Muon's $3\times3$ hidden-group sweep grok, compared with seven of eleven AdamW entries. Their success-conditional mean times are $13{,}011$ and $30{,}486$ steps, a $2.34\times$ ratio. The grids differ and this mean excludes four unsuccessful AdamW configurations. In the selected five-seed runs, Muon's grokking steps span $100$ steps and AdamW's span $7{,}700$. Ratios reach $4.8$--$4.9$ under half width and a smaller modulus. AdamW never groks in the $20\%$ training variant, yielding a horizon-dependent lower bound of $6.6\times$ (Appendix~\ref{app:sweeps}); the depth comparisons are in \S\ref{sec:depth}.

All nine Muon configurations lose test accuracy after grokking; six have a logged evaluation with both train and test accuracy below $95\%$. None is strictly stable: sub-threshold counts range from $2$ to $592$, post-grokking minima reach $0.78\%$, and the counts have no monotone relationship to either hidden hyperparameter. Four of the seven grokking AdamW configurations also collapse, with severity rising monotonically in the learning rate, from none at $10^{-3}$ to $918$ sub-threshold evaluations at $10^{-2}$. Sweeping the readout learning rate over a factor of five and its weight decay from zero to $0.5$ changes the timing and depth of collapse and eliminates none of it; the zero-decay case removes the weight-decay pull on the readout and leaves the failure in place. Across five seeds neither selected setting is strictly stable: Muon records $137$ to $321$ sub-threshold evaluations on every seed with minima to $16.27\%$, and the AdamW reference records one or two on four of five, with a minimum of $27.59\%$. The tested Muon subtraction run also loses test accuracy; its matched AdamW run stays above threshold. Full sweeps, the baseline-selection protocol, and generality variants are in Appendix~\ref{app:sweeps}.

\section{Localizing the collapse at the interface}
\label{sec:localize}

\subsection{Muon updates remain nearly constant at small gradient norms}
\label{sec:quiet}

Figure~\ref{fig:quiet} measures steps $44{,}010$--$44{,}700$ of instrumented continuation I, sampled every update (Appendix~\ref{app:control-provenance}). Its first recorded training failure is at $44{,}702$. Training loss is near $1.5\times10^{-7}$, with hidden and non-hidden gradient norms of order $10^{-6}$ and $10^{-5}$. Muon normalizes its momentum before a fixed Newton--Schulz iteration, reducing sensitivity to its magnitude. Across this window, the log--log slope of the applied hidden step against the hidden gradient norm is $-0.03$. Slopes for the embedding and readout steps against the combined non-hidden norm are about $+1.5$, with $R^2$ across the three fits ranging from $0.84$ to $0.98$. The hidden step and decay largely cancel; their net displacement grows by $64\%$ from step $44{,}100$ to $44{,}700$ while the gradient-driven step changes little. By step $44{,}700$, root-mean-square displacement per parameter from the branch state is $0.073$ for hidden parameters and $0.0091$ for non-hidden parameters, a factor of $8.0$. These measurements establish different displacement scales; their causal interpretation requires the numerical controls in \S\ref{sec:arithmetic}. In the separate ten-step replay used for Figure~\ref{fig:collapse} and Table~\ref{tab:quiet}, training loss rises by $3.8\times10^{8}$ across one interval, and training and test accuracy fall from $100\%$ to $21.12\%$ and $19.04\%$.

\subsection{Which parameters the failure requires}
\label{sec:branches}

\begin{table*}[!t]
\centering
\small
\caption{Two branch suites from the reported step-$44{,}000$ source, each run to step $46{,}000$ and evaluated every ten updates. The group and component suites have separate controls, G and C. This historical configuration uses one shared AdamW group for embeddings and readout (Appendix~\ref{app:control-provenance}). Failure here means sampled training accuracy below $90\%$. Freezing either optimizer-managed side of the interface prevents it over this window; among non-hidden parameters only freezing the unembedding does, and hidden matrices together with the unembedding reproduce it.}
\label{tab:branches}
\begin{tabular}{lrrrr}
\toprule
Branch & First train collapse & Min test & Final test & Max non-hidden grad. \\
\midrule
\multicolumn{5}{l}{\emph{Group branches}} \\
Control G & $44{,}700$ & $16.77\%$ & $100.00\%$ & $913.07$ \\
Freeze hidden & none & $99.07\%$ & $99.08\%$ & $2.79\times10^{-5}$ \\
Freeze non-hidden & none & $100.00\%$ & $100.00\%$ & $0$ \\
\midrule
\multicolumn{5}{l}{\emph{Non-hidden component branches}} \\
Control C & $44{,}710$ & $43.57\%$ & $99.93\%$ & $329.29$ \\
Freeze token embedding & $44{,}910$ & $34.74\%$ & $99.90\%$ & $3278.60$ \\
Freeze position embedding & $44{,}760$ & $37.78\%$ & $99.92\%$ & $323.76$ \\
Freeze unembedding & none & $100.00\%$ & $100.00\%$ & $4.94\times10^{-6}$ \\
Hidden and unembedding only & $45{,}290$ & $40.80\%$ & $100.00\%$ & $348.00$ \\
\bottomrule
\end{tabular}
\end{table*}

Table~\ref{tab:branches} reports two separately executed branch suites, each with its own control and $2{,}000$-step horizon. Their controls have identical logged losses and accuracies at $44{,}000$ and already differ at $44{,}010$. Their earlier source trajectory and the limits of its retained checkpoint and runtime records are described in Appendix~\ref{app:control-provenance}. Freezing either group suppresses the failure for the whole window; freezing the token or position embedding delays it without preventing it; freezing the unembedding prevents it over the same window, and leaving only the hidden matrices and the unembedding free reproduces it. With the non-hidden parameters frozen, Muon continues to apply hidden updates near $0.146$ while loss remains near $1.5\times10^{-7}$.

These branches establish suppression over a finite window; freezing also changes subsequent gradients and optimizer history. To isolate the failing update, Figure~\ref{fig:followup}a swaps features and heads between adjacent checkpoints at five captured events. In every seed, the old head retains or improves accuracy on the updated features, while the new head sharply lowers accuracy on the old features. Fresh linear decoders fitted using only training examples recover $98.20$--$100\%$ at selected collapsed checkpoints (Appendix~\ref{app:basis}). The same protocol gives $0.04$--$0.23\%$ at initialization and $0.02$--$1.12\%$ at sustained memorization, compared with $99.40$--$100\%$ at healthy references (Figure~\ref{fig:followup}b). Decoder selection uses fixed validation indices within each seed, and all $20$ selected refits converge (Appendix~\ref{app:decoder-controls}). These sampled states support retention of information made linearly accessible during training; its emergence between checkpoints remains unresolved. Longer branches still show coadaptation: each decodes its own addition-filtered representation and fails on the other's, and initially identical heads reach cosine similarity $-0.028$. These observations establish readout sensitivity and surviving information. Exact global basis equivalence fails for the measured pairs, despite successful recovery of planted transformations. The joint normalization/orthogonalization ablation also has a separate limitation: non-finite loss was observed in all six generalizing runs (Appendix~\ref{app:ablation}).

\subsection{Freezing embeddings and readout stabilizes the tested depth-1 runs}
\label{sec:prevent}

\begin{figure*}[!t]
\centering
\includegraphics[width=5.062321307011573in]{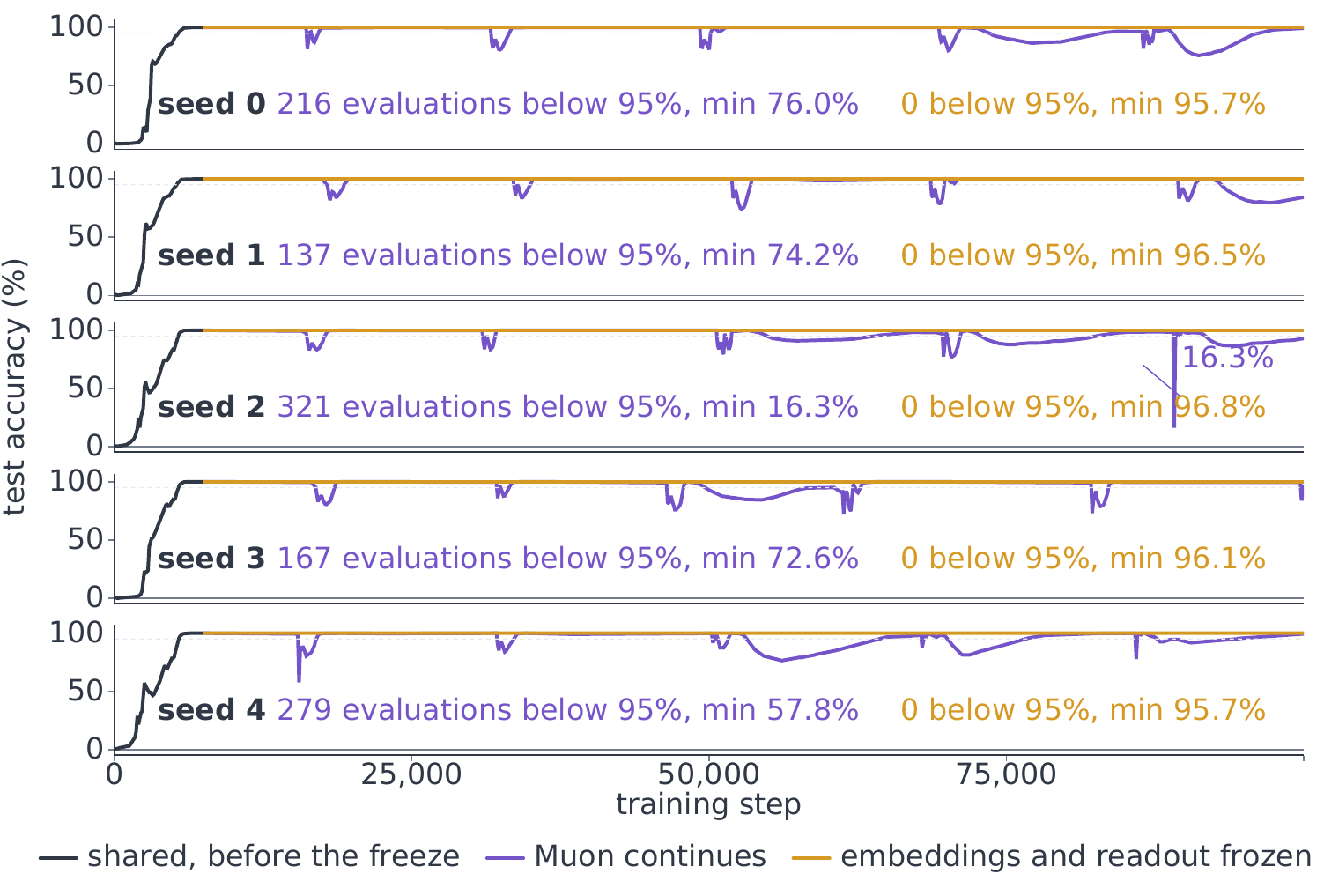}
\caption{The intervention at five seeds. Before the freeze, the paired arms have identical logged metrics. One arm continues under ordinary Muon; the other holds embeddings and unembedding constant. The unfrozen arm falls below threshold $137$ to $321$ times per seed with minima as low as $16.27\%$; the frozen arm records none and never falls below $95.67\%$.}
\label{fig:seeds}
\end{figure*}

We hold the token embedding, the position embedding, and the unembedding constant from roughly two thousand steps after sustained generalization, while hidden matrices continue to receive Muon updates; we call the configuration \emph{Stable Muon}. Across five runs spanning the main condition, $p=97$, training fraction $20\%$, and width $64$, a total of $451{,}400$ post-grokking steps and $4{,}519$ evaluations, no evaluation falls below $95\%$, and in every run the post-grokking minimum is the accuracy at the evaluation that first met the criterion. Figure~\ref{fig:seeds} pairs the arms at five seeds: freezing removes every sub-threshold evaluation on every seed, against an AdamW reference that achieves strict stability on one seed of five. Grokking speed is preserved because the freeze follows the threshold, and per-step cost falls afterward. In the long component comparison, holding only the unembedding fixed leaves eighteen of $936$ evaluations below threshold; also freezing the embeddings removes them (Appendix~\ref{app:prevention}). Arithmetic correction also reaches the original $100{,}000$-step budget (\S\ref{sec:arithmetic}).

\subsection{Numerical controls and the failing step}
\label{sec:arithmetic}

Five matched branches replace stock cross-entropy with a cancellation-resistant \texttt{log1p} evaluation from step $6{,}000$, after grokking and measurable derivative error. The correction keeps model storage, backward matrix arithmetic, and Muon updates in float32. Through step $100{,}000$, the original training horizon, none has a joint train/test failure below $90\%$; minimum measured test accuracy is $98.86\%$, and all finish at $100\%$. Every-update training checks trigger test evaluation below threshold, detecting every joint event. Brief test-only excursions can fall between grid evaluations. 

For each example, zero-sum projection replaces the stock logit derivative $g$ by $g-\mathbf{1}(\mathbf{1}^{\top}g)/p$. From step $15{,}000$ through $20{,}000$, all five projected branches avoid joint failure and finish at $100\%$ test accuracy, although their final derivatives retain $52.94$--$86.07\%$ relative $L_2$ error against a float64 reference at identical float32 logits. Accurate CE and target-derivative repair also give $0/5$ failures over this window. Restoring the class-sum identity therefore prevents the measured failures despite substantial remaining derivative error. At step $15{,}000$, a class-common gradient error produces a class-dependent update difference under the saved Adam state; scalar SGD preserves its class-common structure (Appendix~\ref{app:arithmetic}). These controls support the numerical feature-inflation account of \citet{liu2026glitching}.

From matched step-$6{,}000$ states, reducing Muon's hidden learning rate from $0.03$ to $0.003$, with stock cross-entropy and trainable embeddings and readout, gives $5/5$ joint failures by step $100{,}000$ (Figure~\ref{fig:followup}c). This particular tenfold reduction is insufficient to stabilize the five matched continuations. With unchanged weight-decay coefficients, it also reduces the hidden decoupled weight-decay step (Appendix~\ref{app:lr-reduction}).

\textbf{The mean-dependent head update reproduces the acute failure.} Before the captured failures, full-grid feature-mean norms reach $10{,}374$--$12{,}203$. Appendix~\ref{app:mean-timeline} provides descriptive timelines from the separate archived Muon and AdamW seed study. Let $h_1$ be the post-update residual, $W_{U,0}$ the preceding head, $\Delta W_U$ its actual displacement, and $\mu$ the training-set mean of $h_1$. Starting from the rescued logits $W_{U,0}h_1$, the head-induced change decomposes exactly as
\begin{equation}
\Delta z=\Delta W_U(h_1-\mu)+\Delta W_U\mu.
\label{eq:followup-mean}
\end{equation}
The same training-estimated $\mu$ is used for all test examples. Adding only $\Delta W_U\mu$ nearly reproduces each failure (Table~\ref{tab:mean-update}); its test accuracy differs from the full update by at most $0.33$ percentage points. The centered term retains $87.20$--$100\%$ test accuracy and at least $99.09\%$ training accuracy. This evaluation-only intervention establishes local sufficiency of the mean-dependent term, a class-dependent logit offset shared by all examples. Feature growth and optimizer history remain coupled before the failing update.

\begin{table*}[!t]
\centering\small
\caption{Test accuracy (\%) under the decomposition in Eq.~\ref{eq:followup-mean}. Each partial intervention adds one term to $W_{U,0}h_1$. The full update gives $W_{U,1}h_1$. All means use training examples only.}
\label{tab:mean-update}
\begin{tabular}{lrrrr}
\toprule
Seed & Preceding head & Mean term & Centered term & Full update \\
\midrule
$0$ & $96.39$ & $32.11$ & $95.26$ & $31.87$ \\
$1$ & $94.18$ & $9.42$ & $92.48$ & $9.09$ \\
$2$ & $100.00$ & $0.93$ & $100.00$ & $0.93$ \\
$3$ & $91.01$ & $53.34$ & $87.20$ & $53.60$ \\
$4$ & $98.66$ & $66.07$ & $98.46$ & $66.33$ \\
\bottomrule
\end{tabular}
\end{table*}

Replacing gradients only at the terminal update still fails in all five seeds. Each complete update has negative directional derivative under the accurate training loss, and scales $0.001$ and $0.01$ lower that loss; the full update overshoots (Appendix~\ref{app:mean-update}).

\begin{figure*}[!t]
\centering
\includegraphics[width=6.5in]{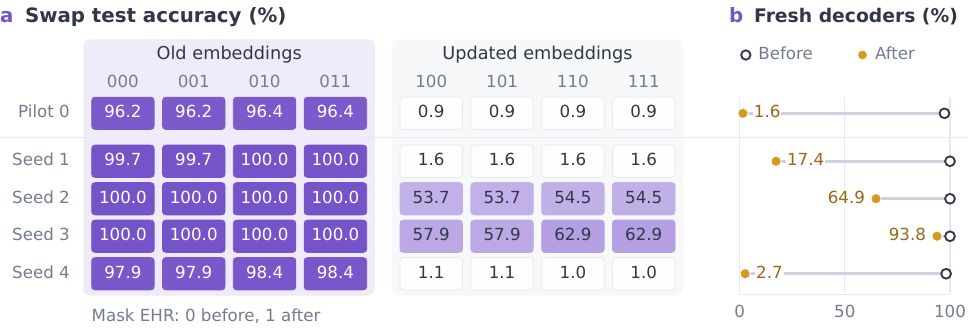}
\caption{Accurate-CE RMS failures: historical CPU pilot $0$ is separated from prospective CPU seeds $1$--$4$. \textbf{(a)} Test accuracy for all eight old/new parameter combinations, ordered embeddings, hidden matrices, readout; $0/1$ denotes before/after the captured update. All old-embedding combinations retain high accuracy; all updated-embedding combinations produce joint train/test failure below $90\%$. \textbf{(b)} Training-only decoders on the input to the readout after final RMS, immediately before/after the same update. All selected refits converge. Protocols, exact counts, replay checks, and sensitivity fits are in Appendix~\ref{app:rms-replication}.}
\label{fig:rms-replication}
\end{figure*}

The corrected subtraction run also remains free of joint failure through $100{,}000$ steps. With gain-free RMS and accurate CE from initialization, all four prospective CPU seeds fail after grokking, at steps $14{,}271$--$15{,}789$. Embedding-only updates reproduce joint failure in each; retaining the old embeddings with updated hidden matrices and readout gives $98.41$--$100\%$ test accuracy. Pre-update derivative errors are below $9\times10^{-8}$ relative. Fresh decoders reach $2.65$--$93.82\%$ after failure, against $98.15$--$100\%$ before it, with all selected refits converged. These results replicate the historical seed-$0$ embedding failure and show variable surviving linear accessibility (Figure~\ref{fig:rms-replication}; Appendix~\ref{app:rms-replication}).

\section{Task-aligned Fourier projections}
\label{sec:family}

We view each cached state as a function of the operand pair. An orthonormal two-dimensional Fourier transform over all $p^2$ pairs partitions its modes into six disjoint families: non-constant \emph{addition} $(k,k)$, \emph{subtraction} $(k,-k)$, $a$-only, $b$-only, the constant mode, and generic interaction. Family \emph{sufficiency} retains that family's modes; \emph{ablation} removes them. Both interventions retain the constant mode when targeting a non-constant family. Unless stated otherwise, Fourier accuracies and solved-checkpoint selection ($\geq95\%$) use the full operand grid. Appendix~\ref{app:fourier} specifies the interventions.

\textbf{The filter supplies task structure.} Retaining all $(k,k)$ modes and the constant mode is exactly the orbit average
\begin{equation}
(P_{\mathrm{add}}h)(a,b)=\frac{1}{p}\sum_{t=0}^{p-1}h(a+t,b-t),
\label{eq:orbit-average}
\end{equation}
with indices modulo $p$. Every donor has the same sum, and the linear readout commutes with this average. For subtraction, the shifts are $(a+t,b+t)$. A lookup-table predictor that outputs the correct one-hot vector on training examples and zero elsewhere becomes perfect after addition filtering whenever every answer orbit contains a training example. Appendix~\ref{app:fourier-control} verifies this control on the experimental splits. Full-family filtered accuracy thus measures the answer-conditioned average of the predictor; high accuracy alone leaves the learned algorithm unidentified.

\begin{table*}[!t]
\centering
\caption{Family sufficiency and ablation on the two operations. Addition columns at depths $2$ and $4$ aggregate eleven solved checkpoints; depth-$1$ columns are matched to the subtraction runs on initialization, configuration, and code path. Chance is $1/113\approx0.885\%$; across forty-three solved checkpoints spanning five seeds and three regimes, the task-aligned family gives exactly $100.00\%$ at every one (Appendix~\ref{app:fourier}).}
\label{tab:swap}
\small
\begin{tabular}{lrrr}
\toprule
 & Addition, depths 2 and 4 & Addition, depth 1 & Subtraction, depth 1 \\
\midrule
$(k,k)$ sufficiency & $100.00\%$ & $100.00\%$ & $0.88\%$ \\
$(k,-k)$ sufficiency & $0.88\%$ & $0.88\%$ & $100.00\%$ \\
$(k,k)$ ablation & $1.34\%$ & $1.51\%$ & $99.68\%$ \\
$(k,-k)$ ablation & $98.26\%$ & $97.51\%$ & $2.54\%$ \\
$a$-only / $b$-only / constant suff. & $0.88\%$ & $0.88\%$ & $0.88\%$ \\
Generic sufficiency & $1.38\%$ & $1.59\%$ & $2.38\%$ \\
Solved checkpoints & $11$ & $6$ & $4$ \\
\bottomrule
\end{tabular}
\end{table*}

\textbf{Task-aligned projection preserves accuracy at the sampled solved checkpoints.} In Table~\ref{tab:swap}, the addition family reaches $100.00\%$ at every selected addition checkpoint. Removing it leaves near-chance accuracy, while removing the subtraction family has little effect. Under matched initialization, configuration, and code path, training on $(a-b)\bmod113$ exchanges these roles: $(k,-k)$ reaches $100\%$ and $(k,k)$ gives chance accuracy. These results characterize the task-aligned projection and family removal; both interventions retain the constant mode. The averaging identity also produces this operation dependence, so identifying shared learned algorithms requires additional evidence \citep{chughtai2023toy}.

\textbf{Frequency and phase matter within the family.} Transporting the family onto other diagonal frequencies at matched power gives $2.64\%$. Randomizing phase per conjugate pair gives $0.19\%$, and randomizing phase per channel gives $0.78\%$. These interventions establish sensitivity to frequency assignment and phase. The memorizer controls in Appendix~\ref{app:fourier-control} also exhibit sparse filtered sufficiency and perturbation sensitivity, so these tests leave algorithm identification unresolved. Two branches diverging from a bit-identical state each decode their own addition-filtered representation at $100\%$ and the other's at chance, showing dependence on the associated readout.

\textbf{The update rule shapes the spread.} On matched runs the Muon representation occupies $326$ effective conjugate pairs of the non-constant spectrum against AdamW's $5$, leaving the task-aligned family holding $28\%$ of non-constant power against AdamW's $91\%$, and the contrast replicates across seeds and inside the family itself (Figure~\ref{fig:grid}; Appendix~\ref{app:dispersion}). These operand-frequency measurements complement prior work on Muon's spectral properties \citep{ruan2026muon,wang2025tailend,wang2026active}. Ablating the normalization and orthogonalization together collapses the count to $4$, past AdamW's, while task-family filtered accuracy remains $100\%$. These comparisons show a change in spectral dispersion. The filtered accuracy retains the averaging interpretation above.

\section{Interference and spectral change}
\label{sec:outvoted}

The task-aligned projection can fail with the model or remain accurate while the full model fails (Figure~\ref{fig:modes}). At a masked checkpoint, full-grid filtered accuracy is $100\%$ and full-model accuracy is $45.85\%$. Rescaling the non-constant task-aligned component raises accuracy to $99.91\%$ without retraining. This is an intervention on a task-informed projection. Appendix~\ref{app:projection-detail} gives the margin decomposition, the scaling curve, and the early AdamW observation of $95.14\%$ full-grid filtered accuracy alongside $0.30\%$ held-out native accuracy. The memorizer control in \S\ref{sec:family} limits what these filtered accuracies establish.

\begin{figure*}[!t]
\centering
\includegraphics[width=5.39in]{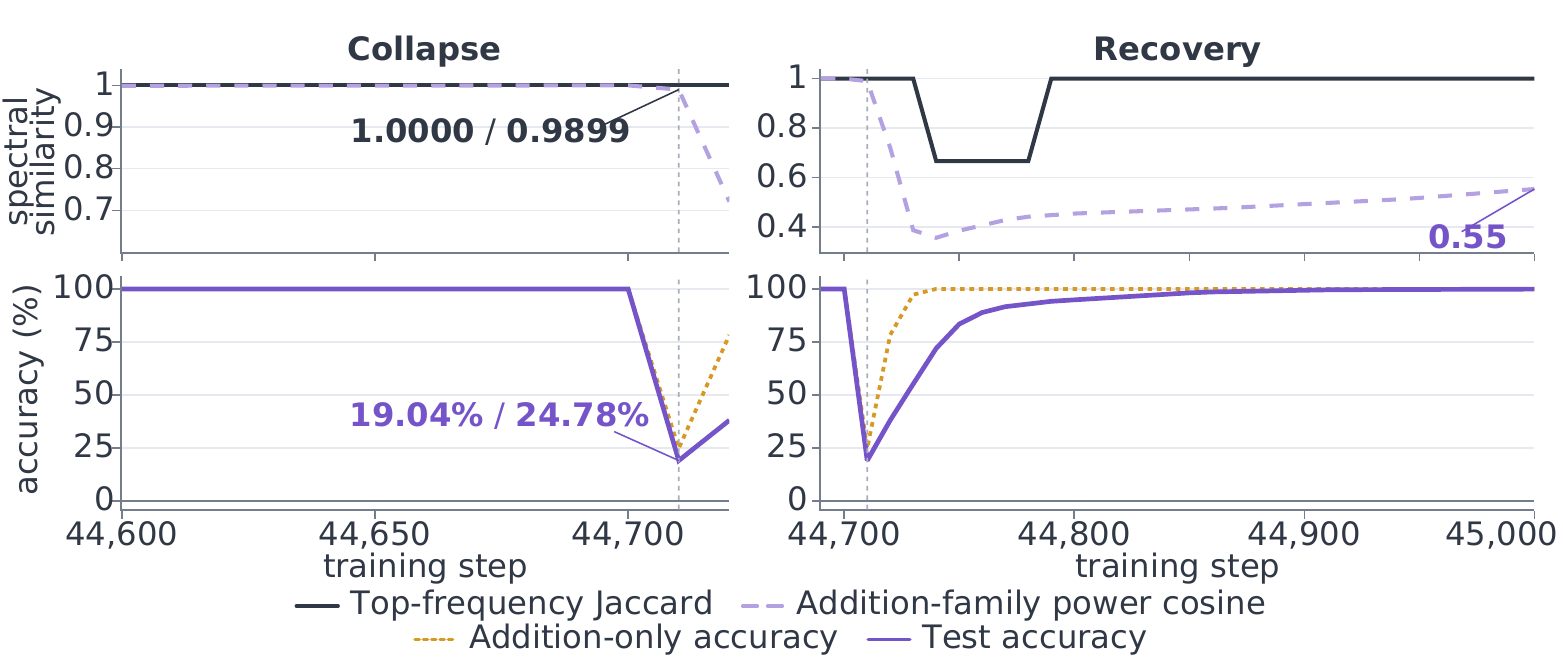}
\caption{Spectral continuation S, evaluated every ten updates through step $45{,}000$. The last sampled healthy state is $44{,}700$ and the first sampled failure is $44{,}710$ (Appendix~\ref{app:control-provenance}). \textbf{Left:} test accuracy falls from $100\%$ to $19.04\%$ and addition-filtered accuracy to $24.78\%$. The five dominant addition-frequency pairs are unchanged (Jaccard $1.0000$), and power over all $56$ addition pairs retains cosine similarity $0.9899$. \textbf{Right:} accuracy returns to $99.90\%$ within $300$ steps, with power cosine $0.55$. The same five pairs remain dominant, with a different allocation of power. The addition family consists of the non-constant $(k,k)$ modes of the final residual's operand-grid Fourier transform; accuracy filters also retain the constant mode.}
\label{fig:collapse}
\end{figure*}

In the ten-step spectral replay, dominant-frequency membership is unchanged across collapse and task-family power retains cosine similarity $0.9899$ (Figure~\ref{fig:collapse}). The filtered margin nevertheless changes from $+18.14$ to $-46.85$. This supports the dissociation between power summaries and predictions. Single-update swaps provide the finer causal localization in Figure~\ref{fig:followup}; exact basis tests and the full spectral analysis are in Appendices~\ref{app:basis} and~\ref{app:projection-detail}.

\section{Depth, stability, and final-readout compatibility}
\label{sec:depth}

At depth $2$, Muon and AdamW reach the grokking criterion at $2{,}300$ and $5{,}100$ steps within a $100{,}000$-step budget ($2.22\times$). At depth $4$, Muon reaches it at $52{,}600$; AdamW remains censored at $300{,}000$, giving a lower bound of $5.7\times$. The AdamW setting selected for stability at depth $1$ records $556$ sub-threshold evaluations at depth $2$. At depth $4$, an exact-state freeze of embeddings and readout at step $126{,}200$ reduces sub-$90\%$ evaluations from $1{,}353$ to $66$ and raises mean post-branch accuracy from $48.07\%$ to $97.50\%$ through step $300{,}000$ ($1{,}738$ post-branch evaluations at $100$-step intervals; Appendix~\ref{app:depth}).

Across $41$ solved checkpoints, intermediate post-block residuals decoded by the final head never exceed $17.97\%$, and only the final block reaches $95\%$. This measures compatibility with that particular head. Appendix~\ref{app:depth} retains the stagewise perturbation analyses, their five-perturbation averages, and the complete depth results.

\section{Related work}
\label{sec:related}

\textbf{Grokking on modular arithmetic.} \citet{power2022grokking} introduced the phenomenon. \citet{nanda2023progress} reverse-engineered a Fourier algorithm, used causal ablations, and introduced restricted and excluded losses to track circuit formation and cleanup. \citet{zhong2023clock} show that hyperparameters select qualitatively different algorithms; \citet{chughtai2023toy} find shared algorithmic structure alongside variation across learned circuits.

\textbf{Numerical instability and adaptive updates.} \citet{thilak2022slingshot} report cyclic instability under adaptive optimizers; \citet{varma2023explaining} describe ungrokking under reduced data. \citet{prieto2025grokking} identify softmax collapse and a loss-minimization direction that scales logits. \citet{liu2026glitching} connect cross-entropy class-sum errors to feature/classifier-mean growth and amplified Adam updates, and test corrections to this numerical feedback. Our matched arithmetic branches extend this evidence to Muon/AdamW routing. Adjacent swaps localize individual failures, and intervention on the mean-dependent part of the actual readout displacement tests its local sufficiency. Joint train/test failure also differs operationally from the perfect-training regime of anti-grokking \citep{prakash2025grokking}.

\citet{wortsman2024smallscale} reproduce attention-logit growth and output-logit divergence in small transformers at high learning rates, and show that independently scaling learning rate and weight decay reduces learning-rate sensitivity. This motivates treating our ordinary LR reduction as a joint change in update size and decay. \citet{molybog2023adam} analyze Adam updates with growing magnitude and weak alignment to descent directions in large-model training. Our captured updates have negative accurate-loss directional derivatives but overshoot at their full scale, providing a measured finite-step qualification for these events.

\textbf{Muon.} \citet{jordan2024muon} introduced the optimizer and \citet{liu2025muon} deployed it at scale under the routing used here. \citet{tveit2025muon} and \citet{wang2026active} report faster grokking in their tested settings; the latter studies orthogonalization and the speed/stability tradeoff with iteration count, including accuracy dips. We examine the five-iteration configuration over longer horizons. \citet{ruan2026muon} and \citet{wang2025tailend} report higher effective rank and a more isotropic weight spectrum under Muon. Our Fourier measurements concern dispersion over operand frequencies, and the joint ablation tests the contribution of normalization and orthogonalization.

\textbf{Representation and readout.} \citet{elhage2021framework} describe residual-stream basis freedom. \citet{anthes2023diagnosing} identify readout misalignment as a major component of forgetting under new tasks. \citet{chou2026twospeeds} use same-split linear probes, geometry, and readout-learning diagnostics to separate representation quality from classifier calibration during grokking and double descent. \citet{gomezjurado2026longdelay} finds a decoder bottleneck in one-step Collatz prediction. Here the task stays fixed: probes at captured post-grokking failures are combined with adjacent-update swaps and a measured-displacement intervention. Initialization and memorization controls calibrate how much linear accessibility precedes the failure.

\section{Discussion}
\label{sec:discussion}

Training-only decoders reach $98.20$--$100\%$, while exact global basis equivalence fails for the measured pairs. Captured updates overshoot accurate local descent directions. Adjacent swaps, the feature-mean intervention, and arithmetic corrections support a numerical contribution to acute readout failure. The relative contributions of feature growth and optimizer history remain unresolved.

The accurate-loss RMS cohort exhibits embedding-localized failures and variable failed-state decodability. The unnormalized controls cover five selected trajectories; the operation extension uses one seed, and the RMS study adds four prospective seeds to a separately identified historical pilot. Behavior beyond $100{,}000$ steps and brief test-only excursions remain unresolved; AdamW collapses limit Muon-specific attribution. Full-family Fourier filtering supplies answer-orbit structure and mixes training and test inputs. The memorizer control rules out inferring an arithmetic algorithm from perfect filtered accuracy alone. Accuracy can collapse through a damaging readout update while useful task information survives, and corrected arithmetic stabilizes the tested unnormalized runs through step $100{,}000$.

\label{sec:main-end}

\clearpage

\subsection*{AI use statement}
Generative AI tools were used for code completion suggestions, manuscript proofreading, and scientific figure preparation. Any generated content was verified, and the authors take full responsibility for the work and the final manuscript.

\bibliography{references}
\bibliographystyle{iclr2026_conference}

\clearpage
\section*{Checklist}

The anonymous repository at \url{https://anonymous.4open.science/r/muon-grokking-BCDC/} provides the code, run logs, raw measurements, retained checkpoints, reproduction commands, and checksum manifests supporting this paper. The study guides explain how to regenerate the analyses and verify the supplied artifacts.

\begin{enumerate}

  \item For all models and algorithms presented, check if you include:
  \begin{enumerate}
    \item A clear description of the mathematical setting, assumptions, algorithm, and/or model. [\textbf{Yes}] Section~\ref{sec:setup} and Appendix~\ref{app:setup} specify the task, architecture, optimizer routing, and update rules. Appendices~\ref{app:followup} and~\ref{app:fourier-control} give the follow-up protocols and projection definitions.
    \item An analysis of the properties and complexity (time, space, sample size) of any algorithm. [\textbf{Yes}] Appendix~\ref{app:computational-cost} gives operation counts and storage bounds for model training, Muon and AdamW updates, Fourier analysis, and decoder fitting. Appendices~\ref{app:setup} and~\ref{app:prevention} report model sizes and measured training throughput.
    \item (Optional) Anonymized source code, with specification of all dependencies, including external libraries. [\textbf{Yes}] The anonymous repository releases the source code. The supplementary source archive includes \nolinkurl{REPRODUCIBILITY.md}, which lists the external dependencies; study protocols record the available software versions and numerical settings.
  \end{enumerate}

  \item For any theoretical claim, check if you include:
  \begin{enumerate}
    \item Statements of the full set of assumptions of all theoretical results. [\textbf{Yes}] The finite operand domain, train/test split, linear readout, and architectural assumptions appear in Sections~\ref{sec:setup} and~\ref{sec:family}. The rank assumptions for basis comparisons and the conditions for the memorizer construction appear in Appendices~\ref{app:basis} and~\ref{app:fourier-control}.
    \item Complete proofs of all theoretical results. [\textbf{Yes}] Section~\ref{sec:arithmetic} derives the readout-displacement decomposition, and Appendix~\ref{app:fourier-control} derives the Fourier and memorizer identities. The repository file \nolinkurl{followup_studies/mathematical_scope.md} gives the constructive basis arguments.
    \item Clear explanations of any assumptions. [\textbf{Yes}] Sections~\ref{sec:setup} and~\ref{sec:family} and Appendices~\ref{app:basis} and~\ref{app:fourier-control} explain where linearity, modulus conditions, rank, and answer-orbit coverage enter the arguments.
  \end{enumerate}

  \item For all figures and tables that present empirical results, check if you include:
  \begin{enumerate}
    \item The code, data, and instructions needed to reproduce the main experimental results (either in the supplemental material or as a URL). [\textbf{Yes}] The anonymous repository contains training and analysis scripts, run logs, raw measurements, retained checkpoints, artifact manifests, and reproduction commands. Appendix~\ref{app:control-provenance} records the limits of exact reconstruction for historical branches.
    \item All the training details (e.g., data splits, hyperparameters, how they were chosen). [\textbf{Yes}] Section~\ref{sec:setup} and Appendices~\ref{app:setup}, \ref{app:sweeps}, and~\ref{app:followup} give splits, optimizer settings, selection criteria, monitoring schedules, and training horizons. The registered follow-up protocols retain the corresponding configurations.
    \item A clear definition of the specific measure or statistics and error bars (e.g., with respect to the random seed after running experiments multiple times). [\textbf{Yes}] Table~\ref{tab:definitions} defines the outcomes and sampling conventions. Figure captions and study protocols identify seed counts, checkpoint selection, perturbation replicates, and the ranges shown by whiskers or shading.
    \item A description of the computing infrastructure used. (e.g., type of GPUs, internal cluster, or cloud provider). [\textbf{Yes}] Appendix~\ref{app:setup} describes the reported original execution backends; Appendix~\ref{app:control-provenance} identifies incomplete historical runtime records. Appendix~\ref{app:rms-replication} specifies the deterministic CPU environment and numerical settings used for the prospective RMS cohort.
  \end{enumerate}

  \item If you are using existing assets (e.g., code, data, models) or curating/releasing new assets, check if you include:
  \begin{enumerate}
    \item Citations of the creator If your work uses existing assets. [\textbf{Yes}] Sections~\ref{sec:intro} and~\ref{sec:related} cite Muon and the source methods used for the architectural, grokking, and numerical analyses. The released code identifies the software libraries it imports.
    \item The license information of the assets, if applicable. [\textbf{Yes}] Appendix~\ref{app:licensing} and the supplementary source archive's \nolinkurl{LICENSING.md} document the applicable third-party licenses, retained notices, and licensing status of the original research assets. The \nolinkurl{licenses/} directory contains the supporting texts and provenance records.
    \item New assets either in the supplemental material or as a URL, if applicable. [\textbf{Yes}] The repository linked above provides code, logs, raw measurements, retained model and optimizer states, decoder fits, and checksum manifests. Reproduction and verification commands accompany the studies.
    \item Information about consent from data providers/curators. [\textbf{Not Applicable}] The examples and labels are generated from finite modular-arithmetic grids.
    \item Discussion of sensible content if applicable, e.g., personally identifiable information or offensive content. [\textbf{Not Applicable}] The experimental data consist of integer operands and modular-arithmetic labels.
  \end{enumerate}

  \item If you used crowdsourcing or conducted research with human subjects, check if you include:
  \begin{enumerate}
    \item The full text of instructions given to participants and screenshots. [\textbf{Not Applicable}] The study uses no crowdsourcing or human participants.
    \item Descriptions of potential participant risks, with links to Institutional Review Board (IRB) approvals if applicable. [\textbf{Not Applicable}] The study uses no human participants.
    \item The estimated hourly wage paid to participants and the total amount spent on participant compensation. [\textbf{Not Applicable}] The study uses no paid participants.
  \end{enumerate}

\end{enumerate}

\clearpage
\appendix
\onecolumn
\section{Setup details and operational definitions}
\label{app:setup}

\subsubsection*{Reproducibility statement}
The original experiments run on a single device from released code. The anonymized repository at \url{https://anonymous.4open.science/r/muon-grokking-BCDC/} contains the training and analysis scripts, $191$ CSV run logs, and $20$ HTML summaries. Follow-up artifacts add raw diagnostics, model/optimizer/random-state checkpoints, replay checks, and the scripts for every new table and figure. Shared-state checks cover the matched depth-$4$ freeze and fresh arithmetic and learning-rate continuations; Appendix~\ref{app:control-provenance} documents historical branch provenance. Appendix~\ref{app:setup} specifies the original setup; Appendix~\ref{app:followup} gives the decoder, basis, arithmetic, and architecture-control protocols and their fixed horizons.

\begin{table}[!htbp]
\centering\small
\caption{Operational definitions. The original training grid is $100$ updates. Five consecutive scheduled evaluations span $400$ updates; follow-up confirmation uses six and spans $500$. Finer event-capture schedules are identified by trajectory.}
\label{tab:definitions}
\begin{tabular}{p{0.25\textwidth}p{0.68\textwidth}}
\toprule
Term & Definition \\
\midrule
Sustained $99.9\%$ train & First of five consecutive scheduled training evaluations at or above $99.9\%$. \\
Original grokking & First of five consecutive scheduled test evaluations at or above $95\%$. Used for the original speed comparisons. \\
Memorization plateau & Interval from sustained-$99.9\%$ training accuracy to original grokking; its duration is the difference between these steps. \\
Follow-up confirmation & Sixth consecutive scheduled test evaluation at or above $95\%$. Prospective joint-event monitoring begins after this confirmation. \\
Strict stability & Every subsequent sampled test accuracy is at least $95\%$ over the stated horizon. \\
Joint failure & Training and held-out test accuracy are both below $90\%$ at the same state, after confirmation where required by the protocol. \\
Table~\ref{tab:branches} failure & Training accuracy below $90\%$ at a scheduled branch evaluation. This criterion does not require a simultaneous test threshold. \\
First detected failure & Earliest qualifying observation under the stated monitoring schedule. It can follow the underlying threshold crossing when sampling is sparse. \\
First captured failure & Earliest qualifying saved state under the specified capture protocol. Its window and resolution can differ from the original detection schedule. \\
Minimum sampled accuracy & Minimum over recorded evaluations in the stated window, with the inclusion of the source or confirmation state specified. \\
Final accuracy & Accuracy at the reported terminal state; recovery can make it higher than the post-grokking minimum. \\
Evaluation count & Number of recorded evaluations satisfying the stated threshold, distinguishing scheduled and training-triggered evaluations. \\
Seed-level incidence & Number of seeds with at least one qualifying event divided by the number of evaluable seeds over the stated horizons. Repeated events within a seed count once. \\
\bottomrule
\end{tabular}
\end{table}

\begin{table}[t]
\centering\small
\caption{Selected configurations. Muon and Stable Muon share the configuration before freezing; the paired CPU runs have identical logged pre-freeze metrics. Historical branch provenance is given in Appendix~\ref{app:control-provenance}.}
\label{tab:configs}
\begin{tabular}{llll}
\toprule
Configuration & Group & Optimizer & Settings \\
\midrule
AdamW baseline & All parameters & AdamW & $\text{lr}=10^{-3}$, $\text{wd}=3.0$ \\
\midrule
\multirow{3}{*}{Muon} & Hidden & Muon & $\text{lr}=0.03$, $\text{wd}=0.1$, $\beta=0.95$, $\text{NS}=5$ \\
 & Auxiliary & AdamW & $\text{lr}=10^{-3}$, $\text{wd}=1.0$ \\
 & Readout & AdamW & $\text{lr}=2.5\times10^{-4}$, $\text{wd}=1.0$ \\
\midrule
Stable Muon & \multicolumn{3}{l}{Muon, with auxiliary and readout groups frozen after sustained generalization} \\
\bottomrule
\end{tabular}
\end{table}

Each example is the token sequence $[a,b,=]$; the vocabulary contains $p+1=114$ tokens and the output layer $p=113$ classes. The ordered operand grid contains $p^2=12{,}769$ examples; a fixed random $30\%$ (unless stated) forms the training set, generated once from the run seed and shared by all matched comparisons. The original sweeps load saved initial states, evaluate every $100$ steps, and checkpoint every $1{,}000$. Their budgets are $100{,}000$ steps at depths 1 and 2 and $300{,}000$ at depth 4. Diagnostic branches use the shorter horizons and finer grids specified below.

Each block holds four two-dimensional matrices (fused QKV $384\times128$, attention output $128\times128$, MLP input $512\times128$, MLP output $128\times512$), $196{,}608$ parameters; with embeddings and unembedding the depth-1 model has $226{,}048$ parameters. Depth is varied by appending blocks under nested initialization, so deeper models begin from a strict superset of the shallower model's initial state. Muon maintains an exponential moving average of the gradient with $\beta=0.95$, forms a Nesterov-style update, divides it by its Frobenius norm plus $10^{-7}$, applies five quintic Newton--Schulz iterations with coefficients $(3.4445,-4.7750,2.0315)$, scales by $\sqrt{\max(1,\text{rows}/\text{cols})}$, and applies decoupled weight decay; the implementation refuses any parameter that is not a matrix. All AdamW groups use $\beta_1=0.9$, $\beta_2=0.999$.

Table~\ref{tab:definitions} distinguishes the original and follow-up definitions. Original scheduled evaluations occur every $100$ steps; their five-check criterion spans $400$ steps, while follow-up confirmation occurs on the sixth qualifying check. The original study reports accelerator execution for the single-seed sweeps and CPU execution for the five-seed replications. Historical runtime manifests are incomplete (Appendix~\ref{app:control-provenance}). The Muon and Stable Muon CPU arms have identical logged metrics over the seventy-four to seventy-five evaluations preceding the freeze. This check compares recorded metrics; it does not compare every parameter tensor at each evaluation. Fresh follow-ups use the explicitly recorded deterministic CPU configuration and verify the replays described in Appendix~\ref{app:followup}. The seed sets the initialization and the train/test split together, so the five seeds are five draws of the problem instance.

\subsection{Historical controls and event identities}
\label{app:control-provenance}

Table~\ref{tab:branches} contains two separately executed control continuations. Control G belongs to the parameter-group suite, and Control C to the auxiliary-component suite. Both report the same losses and accuracies at step $44{,}000$, matching the archived hidden-LR-$0.01$, seed-$0$ trajectory: training loss $1.86781946\times10^{-7}$, test loss $3.32467862\times10^{-5}$, and $100\%$ accuracy on both splits. Their logged losses already differ at $44{,}010$. Each suite is evaluated every ten updates through $46{,}000$, giving $201$ samples including the source state. The reported minimum test accuracies, $16.77\%$ and $43.57\%$, reproduce the corresponding CSV minima; final accuracies and maximum auxiliary-gradient norms likewise reproduce the last row and column maxima. The first-failure column uses training accuracy below $90\%$ after memorization.

The historical branch programs load the model, Muon, and auxiliary AdamW state dictionaries. This earlier design uses one AdamW group for embeddings and readout. They reseed Python and PyTorch from the saved seed, including CUDA when available; their checkpoint format omits saved RNG states and backend metadata. The model is full-batch and has no dropout. The original step-$44{,}000$ checkpoint tensors, launch commands, and runtime manifests are unavailable in the retained artifacts. Matching initial metrics therefore cannot certify bitwise equality of the two restored states or identify the original backend. The logged results remain verified; the source of their subsequent divergence is unresolved.

We separately test implementation parity on an available seed-$0$, step-$16{,}000$ checkpoint. To match the earlier optimizer layout, we merge the head into the auxiliary AdamW group at learning rate $0.001$, preserving all saved moments and step counters. The historical group, component, and instrumented group programs each run $2{,}000$ updates on CPU with PyTorch $2.13.0$, one thread, and deterministic algorithms. All $201$ shared metric rows agree exactly, and terminal model, optimizer, and PyTorch RNG states are bitwise equal. This replay verifies agreement of the three implementations under the registered CPU test; its adapted source state is separately identified in the artifacts.

\begin{table}[htbp]
\centering\small
\caption{Distinct historical continuations around step $44{,}700$. Accuracies are percentages at each run's first observed training failure. The spectral comparison uses S's healthy $44{,}700$ state and failing $44{,}710$ state.}
\label{tab:event-provenance}
\begin{tabular}{llrrrr}
\toprule
ID & Analysis & Sample interval & End step & Failure step & Train / test \\
\midrule
G & Table~\ref{tab:branches}, groups & $10$ & $46{,}000$ & $44{,}700$ & $18.54 / 16.77$ \\
C & Table~\ref{tab:branches}, components & $10$ & $46{,}000$ & $44{,}710$ & $47.70 / 43.57$ \\
I & Applied-step diagnostics & $1$ & $46{,}000$ & $44{,}702$ & $24.31 / 22.12$ \\
S & Spectral replay & $10$ & $45{,}000$ & $44{,}710$ & $21.12 / 19.04$ \\
\bottomrule
\end{tabular}
\end{table}

The source files under \texttt{runs/} are \nolinkurl{branch_control_from_44000.csv} (G), \nolinkurl{aux_control_from_44000.csv} (C), \nolinkurl{branch_control_from_44000_instrumented.csv} (I), and \nolinkurl{collapse_spectral_replay_from_44000.csv} (S). Their sampling intervals and terminal horizons identify separate trajectories even when event timestamps coincide. The ten-update grids can miss intervening excursions. Source hashes, exact summary calculations, and the controlled replay protocol accompany the paper.

\section{Sweeps, baseline selection, and generality}
\label{app:sweeps}

\begin{table}[t]
\centering\small
\caption{AdamW sweep. Seven of eleven retained configurations reach sustained $95\%$ test
accuracy. Post-grokking evaluations below $95\%$ increase monotonically with the learning
rate.}
\label{tab:adamw-sweep}
\begin{tabular}{llrrrl}
\toprule
LR & WD & Sustained $95\%$ & Min after grokking & Evals $<95\%$ & Strictly stable \\
\midrule
$3\times10^{-4}$ & $0.1$ & no grok & & & \\
$3\times10^{-4}$ & $1.0$ & no grok & & & \\
$3\times10^{-4}$ & $3.0$ & $25{,}500$ & $95.67\%$ & $0$ & yes \\
$10^{-3}$ & $0.1$ & no grok & & & \\
$10^{-3}$ & $1.0$ & $52{,}500$ & $96.49\%$ & $0$ & yes \\
$10^{-3}$ & $3.0$ & $8{,}200$ & $95.01\%$ & $0$ & yes \\
$3\times10^{-3}$ & $0.1$ & $49{,}500$ & $13.26\%$ & $2$ & no \\
$3\times10^{-3}$ & $1.0$ & $8{,}500$ & $53.72\%$ & $4$ & no \\
$10^{-2}$ & $0.1$ & $62{,}900$ & $0.54\%$ & $258$ & no \\
$10^{-2}$ & $1.0$ & $6{,}300$ & $0.41\%$ & $918$ & no \\
$10^{-2}$ & $3.0$ & no grok & & & \\
\bottomrule
\end{tabular}
\end{table}

\begin{table}[t]
\centering\small
\caption{Muon sweep over the hidden group. All nine configurations reach sustained $95\%$
test accuracy and none is strictly stable. The count of sub-threshold evaluations has no
monotone relationship to either hyperparameter.}
\label{tab:muon-sweep}
\begin{tabular}{llrrrl}
\toprule
LR & WD & Sustained $95\%$ & Min after grokking & Evals $<95\%$ & Strictly stable \\
\midrule
$0.03$ & $0.1$ & $5{,}400$ & $69.40\%$ & $56$ & no \\
$0.01$ & $0.1$ & $6{,}000$ & $53.24\%$ & $107$ & no \\
$0.03$ & $0.3$ & $7{,}200$ & $75.63\%$ & $2$ & no \\
$0.01$ & $0.3$ & $8{,}500$ & $76.60\%$ & $145$ & no \\
$0.03$ & $0.03$ & $9{,}000$ & $0.78\%$ & $592$ & no \\
$0.003$ & $0.1$ & $14{,}400$ & $90.07\%$ & $5$ & no \\
$0.003$ & $0.03$ & $19{,}000$ & $21.24\%$ & $130$ & no \\
$0.01$ & $0.03$ & $19{,}500$ & $0.83\%$ & $473$ & no \\
$0.003$ & $0.3$ & $28{,}100$ & $78.59\%$ & $265$ & no \\
\bottomrule
\end{tabular}
\end{table}

\begin{table}[t]
\centering\small
\caption{Controlled variation of modulus, training fraction, and width. Stable Muon and
Muon share a configuration up to the freeze, so differences in their grokking step measure
pre-intervention run-to-run variation on this backend.}
\label{tab:generality}
\begin{tabular}{llrrl}
\toprule
Variant & Regime & Sustained $95\%$ & Min post-grokking & Outcome \\
\midrule
\multirow{3}{*}{$p=97$, train $30\%$, width $128$}
 & AdamW & $31{,}000$ & $95.52\%$ & stable \\
 & Muon & $6{,}300$ & $15.49\%$ & unstable \\
 & Stable Muon & $6{,}300$ & $95.05\%$ & stable \\
\midrule
\multirow{3}{*}{$p=113$, train $20\%$, width $128$}
 & AdamW & no grok & & \\
 & Muon & $15{,}100$ & $49.25\%$ & unstable \\
 & Stable Muon & $14{,}500$ & $95.16\%$ & stable \\
\midrule
\multirow{3}{*}{$p=113$, train $30\%$, width $64$}
 & AdamW & $84{,}800$ & $96.38\%$ & stable \\
 & Muon & $17{,}700$ & $53.23\%$ & unstable \\
 & Stable Muon & $16{,}500$ & $95.67\%$ & stable \\
\bottomrule
\end{tabular}
\end{table}

Tables~\ref{tab:adamw-sweep} and~\ref{tab:muon-sweep} give both sweeps under identical architecture, split, and budget. The fastest AdamW configuration ($\mathrm{lr}=10^{-2}$, $\mathrm{wd}=1.0$) is unusable: $918$ sub-threshold evaluations at this seed, and across seeds three of five never reach sustained $95\%$, the two that do record $805$ and $956$, and all five end near chance. The baseline is therefore the fastest configuration that is strictly stable in this sweep, $\mathrm{lr}=10^{-3}$, $\mathrm{wd}=3.0$; the seed replication shows strict stability at that setting is a property of one seed rather than of the configuration. For Muon no configuration is strictly stable, so the selected configuration is the fastest.

Both optimizers fit the training set between steps $200$ and $800$ in every grokking configuration, so the difference lies almost entirely in the delay from memorization to generalization: $30{,}100$ steps for AdamW against $12{,}544$ for Muon averaged over successes. Across five seeds Muon reaches sustained generalization between $5{,}300$ and $5{,}400$ steps and the AdamW reference between $4{,}000$ and $11{,}700$, medians $5{,}400$ and $8{,}300$.

Varying the readout learning rate over $10^{-4}$, $2.5\times10^{-4}$, and $5\times10^{-4}$ gives post-grokking minima of $37.13\%$, $79.40\%$, and $3.81\%$; varying its weight decay over $0$, $0.1$, and $0.5$ gives $33.34\%$, $0.79\%$, and $1.12\%$; every setting is unstable. On $(a-b)\bmod 113$ the Muon run records $18$ post-grokking evaluations below $95\%$ with a minimum of $1.52\%$ and the AdamW run records none, reproducing the depth-1 addition asymmetry at $194$ and none. Table~\ref{tab:generality} varies modulus, training fraction, and width; Stable Muon and Muon share a configuration up to the freeze, so differences in their grokking steps bound run-to-run variation on this backend.

\section{The quiet window}
\label{app:quiet}

\begin{figure}[t]
\centering
\includegraphics[width=\textwidth]{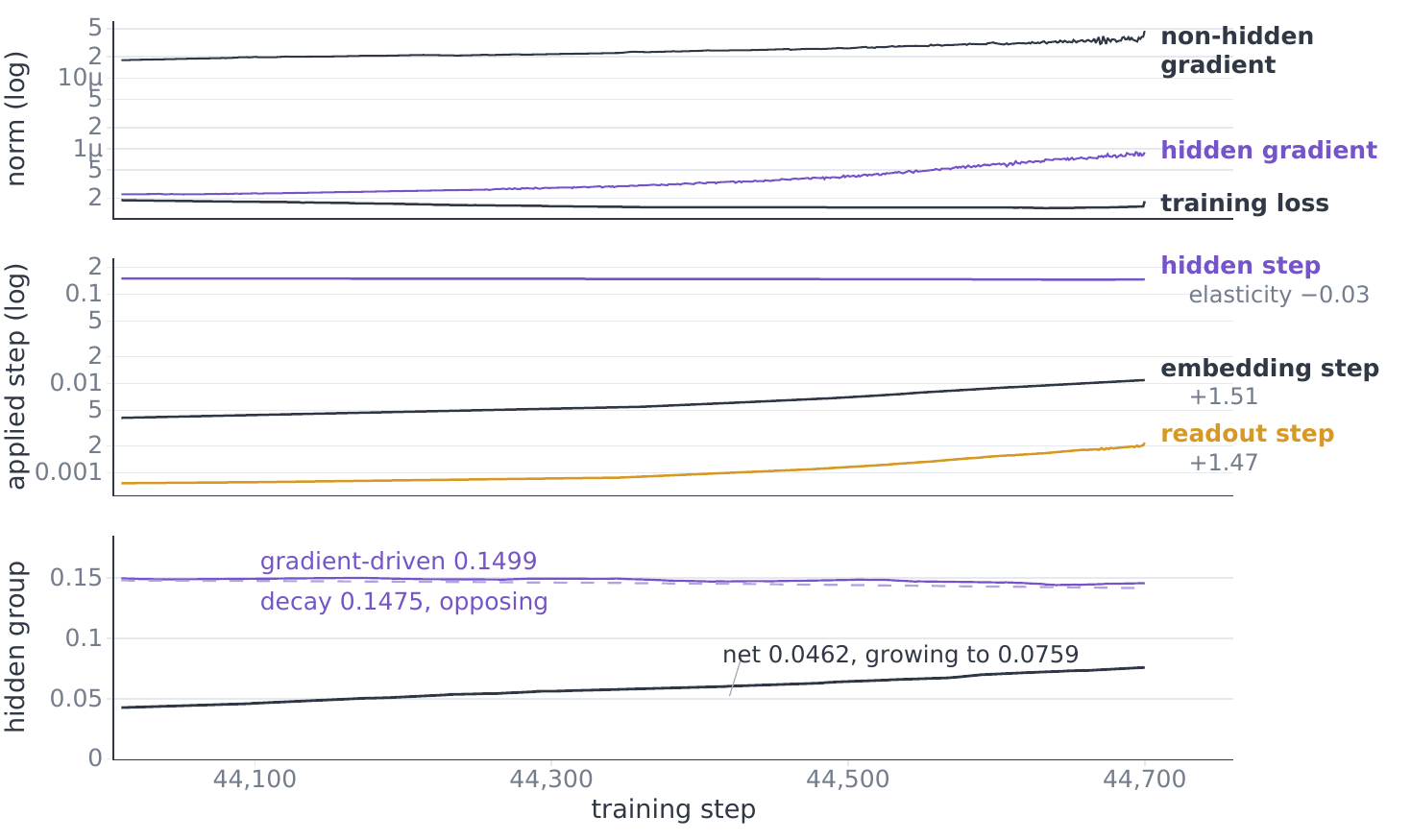}
\caption{Every-step measurements over $691$ updates before a collapse. \textbf{Top:} loss and gradient norms in the quiet window. \textbf{Middle:} gradient-driven applied steps and log--log slopes against the hidden gradient norm for Muon and the combined non-hidden gradient norm for the embedding and readout parameter subsets. \textbf{Bottom:} hidden step and decay magnitudes differ by under $2\%$ at step $44{,}100$. Their vector sum is smaller and grows by $64\%$ from step $44{,}100$ to $44{,}700$ while the gradient-driven step changes little.}
\label{fig:quiet}
\end{figure}

\begin{table}[t]
\centering\small
\caption{Diagnostics through the quiet window and across the collapse. The training loss
and both gradient norms are numerically negligible until the failure. The Muon update column
is the orthogonalized step before weight decay, varying by under three percent across
the pre-collapse rows. These rows come from spectral continuation S, sampled every ten updates. Net-step diagnostics use continuation I, sampled every update, whose first training failure is $44{,}702$ (Appendix~\ref{app:control-provenance}).}
\label{tab:quiet}
\begin{tabular}{lrrrr}
\toprule
Step & Train loss & Hidden grad.\ norm & Non-hidden grad.\ norm & Muon update norm \\
\midrule
$44{,}100$ & $1.77\times10^{-7}$ & $2.33\times10^{-7}$ & $1.95\times10^{-5}$ & $0.1498$ \\
$44{,}300$ & $1.55\times10^{-7}$ & $2.73\times10^{-7}$ & $2.20\times10^{-5}$ & $0.1494$ \\
$44{,}500$ & $1.50\times10^{-7}$ & $4.06\times10^{-7}$ & $2.67\times10^{-5}$ & $0.1481$ \\
$44{,}700$ & $1.53\times10^{-7}$ & $8.64\times10^{-7}$ & $3.52\times10^{-5}$ & $0.1455$ \\
\midrule
$44{,}710$ & $57.50$ & $25.26$ & $836.10$ & $0.1429$ \\
$44{,}720$ & $6.58$ & $5.09$ & $92.55$ & $0.1655$ \\
\bottomrule
\end{tabular}
\end{table}

Figure~\ref{fig:quiet} uses the every-step instrumented rerun, with a quiet window from $44{,}010$ through $44{,}700$ and failure at $44{,}702$. Table~\ref{tab:quiet} uses a separate ten-step spectral replay, whose first observed failure is at $44{,}710$. The following step/decay diagnostics use the instrumented rerun. On the hidden group the gradient-driven component is $0.1499$ at step $44{,}100$ against a decay component of $0.1475$, a ratio of $0.984$, leaving a net displacement of $0.0462$ that grows to $0.0759$ across the window while the orthogonalized step stays flat; over the window the hidden gradient norm grows by $3.92$ while its gradient-driven step changes by $0.97$, and the non-hidden gradient grows by $2.60$ while the embedding and readout steps grow by $2.65$ and $2.84$. The readout decay component has larger norm than its gradient-driven component on $14.9\%$ of quiet-window steps. In the ten-step spectral replay, the combined non-hidden gradient norm rises from $3.52\times10^{-5}$ to $836.10$; the orthogonalized Muon step over that interval is $0.1429$, lower than the $0.1455$ preceding it, and its maximum over the branch occurs after the event.

\section{Ablating the normalized-orthogonalization path}
\label{app:ablation}

\begin{table}[t]
\centering\small
\caption{Ablated runs that reach sustained generalization. Minimum is the lowest test
accuracy from the sustained-$95$ step to the last recorded evaluation. Terminal is the step
at which non-finite loss was observed during the original run.}
\label{tab:nons}
\begin{tabular}{llrrrr}
\toprule
Learning rate & Backend & Precision & Grokked & Minimum & Terminal \\
\midrule
$0.3$ & MPS & \texttt{float32} & $3{,}100$ & $97.28\%$ & $23{,}514$ \\
$0.3$ & CPU & \texttt{float32} & $3{,}100$ & $97.52\%$ & $24{,}485$ \\
$0.3$ & CPU & \texttt{float64} & $3{,}100$ & $97.06\%$ & $22{,}889$ \\
$1.0$ & MPS & \texttt{float32} & $6{,}800$ & $97.11\%$ & $23{,}800$ \\
$1.0$ & CPU & \texttt{float64} & $7{,}700$ & $95.96\%$ & $29{,}831$ \\
$1.0$ & CPU & \texttt{float32} & $6{,}500$ & $95.60\%$ & $38{,}955$ \\
\bottomrule
\end{tabular}
\end{table}

The ablation removes Frobenius normalization and the Newton--Schulz iteration together, retaining momentum, shape scaling, and weight decay. In the tested sweep, learning requires a rate of $0.3$ or $1.0$. Six ablated runs generalize, with no recorded post-grokking evaluation below $95\%$ (Table~\ref{tab:nons}). All six were observed to terminate with non-finite loss, in float32 and float64, at steps $22{,}889$--$38{,}955$. The matched Muon comparator records thirteen sub-threshold evaluations through step $22{,}800$. Over its full $100{,}000$ steps, its minimum and final accuracies are $29.49\%$ and $96.48\%$. At the analyzed checkpoints, ablation reduces dispersion from $326.09$ effective conjugate pairs to $4.11$, while the task-aligned family retains $100\%$ sufficiency (Appendix~\ref{app:dispersion}).

\section{Prevention details}
\label{app:prevention}

\begin{table}[t]
\centering\small
\caption{Stable Muon across five runs covering four settings; the main condition appears
twice, under a fixed freeze step and under the automatic trigger. The minimum is the lowest
test accuracy at
any evaluation from the sustained-$95$ step onward, and in every run it is the value at
that step itself.}
\label{tab:stable}
\begin{tabular}{lrrrrr}
\toprule
Run & Grokked & \shortstack{First frozen\\eval.} & Post-grokking steps & Minimum & Evals $<95\%$ \\
\midrule
Main condition & $5{,}900$ & $8{,}100$ & $94{,}100$ & $95.34\%$ & $0$ \\
Selected Stable Muon & $5{,}400$ & $7{,}500$ & $94{,}600$ & $95.74\%$ & $0$ \\
$p = 97$ & $6{,}300$ & $8{,}400$ & $93{,}700$ & $95.05\%$ & $0$ \\
Training fraction $20\%$ & $14{,}500$ & $16{,}600$ & $85{,}500$ & $95.16\%$ & $0$ \\
Width $64$ & $16{,}500$ & $18{,}600$ & $83{,}500$ & $95.67\%$ & $0$ \\
\midrule
Total & & & $451{,}400$ & & $0$ \\
\bottomrule
\end{tabular}
\end{table}

In four of the five runs the freeze is triggered automatically, $2{,}000$ steps after the first evaluation of a five-evaluation streak at or above $95\%$; the main-condition run uses a fixed step. The first logged frozen evaluation lands $2{,}100$ to $2{,}200$ steps after sustained generalization; the actual freezes occur $100$ steps earlier. Within a depth-2 run the median step costs $57.87$\,ms before the freeze and $54.72$ after; at depth 4, $111.39$ and $107.75$; pre-freeze rates match ordinary Muon's. Holding only the unembedding constant from the same step leaves the run unstable, minimum $83.13\%$ and eighteen of $936$ evaluations below $95\%$; freezing the embeddings and unembedding together at the same step produces none.

\section{Fourier decomposition and intervention suite}
\label{app:fourier}

\begin{table}[t]
\centering\small
\caption{The six mode families. The families are pairwise disjoint and exhaust the
$p \times p$ grid, so any statement about one family and its complement is an exact
accounting of the whole representation.}
\label{tab:families}
\begin{tabular}{lll}
\toprule
Family & Modes & Count at $p = 113$ \\
\midrule
Addition & $(k, k)$, $k \neq 0$ & $112$ \\
Subtraction & $(k, -k)$, $k \neq 0$ & $112$ \\
$a$-only & $(k, 0)$, $k \neq 0$ & $112$ \\
$b$-only & $(0, l)$, $l \neq 0$ & $112$ \\
Constant & $(0, 0)$ & $1$ \\
Generic interaction & all remaining $(k, l)$ & $12{,}320$ \\
\midrule
Total & & $12{,}769$ \\
\bottomrule
\end{tabular}
\end{table}

\begin{table}[t]
\centering\small
\setlength{\tabcolsep}{5pt}
\caption{Interventions applied to a cached state. Preserved and altered quantities specify the
intervention; an accuracy change measures sensitivity to that perturbation. Non-constant-family
sufficiency and ablation both retain the constant mode.}
\label{tab:interventions}
\begin{tabular}{lll}
\toprule
Intervention & Preserved & Destroyed \\
\midrule
Family sufficiency & Selected family and constant mode & Other non-constant modes \\
Family ablation & All other families & Modes in one family \\
Equal-power relocation & Total power, family membership & Which frequencies carry it \\
Random native subset & Pair count, native coefficients & Which of the used frequencies survive \\
Pair-global phase scramble & Support and power exactly & Relative phase across pairs \\
Channelwise phase scramble & Support and power exactly & Relative phase across channels \\
Cross-readout substitution & Both representations intact & Pairing of state with its own readout \\
\bottomrule
\end{tabular}
\end{table}

\begin{table}[t]
\centering\small
\caption{Family interventions at forty-three solved checkpoints, spanning five seeds and the
Muon, Stable Muon, and AdamW regimes. Sufficiency retains the named family and constant mode; ablation
removes the named non-constant family and retains the rest. Accuracies and selection use the full grid. The chance rate is $0.88\%$.}
\label{tab:families-result}
\begin{tabular}{lrrrr}
\toprule
Family & Sufficiency & Range & Ablation & Range \\
\midrule
Addition & $100.00\%$ & $[100.00,\,100.00]$ & $1.71\%$ & $[0.67,\,4.24]$ \\
Subtraction & $0.88\%$ & $[0.88,\,0.88]$ & $99.61\%$ & $[95.79,\,100.00]$ \\
$a$-only & $0.88\%$ & $[0.88,\,0.88]$ & $99.85\%$ & $[97.76,\,100.00]$ \\
$b$-only & $0.88\%$ & $[0.88,\,0.88]$ & $99.85\%$ & $[97.72,\,100.00]$ \\
Constant & $0.88\%$ & $[0.88,\,0.88]$ & & \\
Generic interaction & $3.28\%$ & $[0.78,\,14.21]$ & $99.99\%$ & $[99.78,\,100.00]$ \\
\bottomrule
\end{tabular}
\end{table}

For a chosen point in the forward pass the model runs on all $p^2$ operand pairs, the states form a tensor $X\in\mathbb{R}^{p\times p\times d}$, and an orthonormal two-dimensional DFT over the operand axes gives $\widehat{X}[k,l,:]$ with power $\lVert\widehat{X}[k,l,:]\rVert^2$. Because $X$ is real, modes occur in conjugate pairs and every filter retains or removes both members together; transforms run in single precision on CPU. The families of Table~\ref{tab:families} are pairwise disjoint and exhaust the grid, and addition and subtraction are disjoint because $p$ is odd. Power fractions are normalized against non-constant power; every non-constant-family sufficiency filter also retains the constant mode, which becomes a fixed logit offset under the bias-free unembedding. Ablation removes only the selected non-constant modes, retaining this offset. The two masks are complementary on the non-constant spectrum and share the constant mode. In the margin decomposition and $\alpha$ sweep, the task-aligned component excludes the constant mode, which belongs to the remainder. Solved-checkpoint selection and intervention accuracies use the full operand grid; training trajectories explicitly labeled test accuracy use the held-out split.

Table~\ref{tab:families-result} applies sufficiency and ablation to each family at forty-three solved checkpoints spanning five seeds and the Muon, Stable Muon, and AdamW regimes. Equal-power relocation ranges $0.88\%$ to $15.04\%$ over $430$ replicates. Among the tested descending-power prefixes, the smallest reaching $95\%$ contains two to four conjugate pairs at AdamW checkpoints (median $3$), against three to seventeen at Muon (median $12.5$) and eight to nineteen at Stable Muon (median $11$). Along the AdamW trajectory of \S\ref{sec:outvoted}, the five strongest pairs give $4.91\%$ at step $3{,}000$, $19.48\%$ at $20{,}000$, $70.73\%$ at $30{,}000$, and $100\%$ at $37{,}000$, while the family's share rises from $1.05\%$ through $1.66\%$ and $11.08\%$ to $70.62\%$, reaching $90.51\%$ by step $100{,}000$; test accuracy over those steps is $0.30\%$, $0.29\%$, $2.08\%$, and $100\%$.

The full-family accuracies above include the orbit average in Eq.~\ref{eq:orbit-average}. Prefix filtering and phase or frequency perturbations measure additional spectral structure, but their interpretation as evidence of a learned algorithm requires controls against memorization. Appendix~\ref{app:fourier-control} supplies that comparison.

\section{Spectral dispersion}
\label{app:dispersion}

\begin{figure}[t]
\centering
\includegraphics[width=\textwidth]{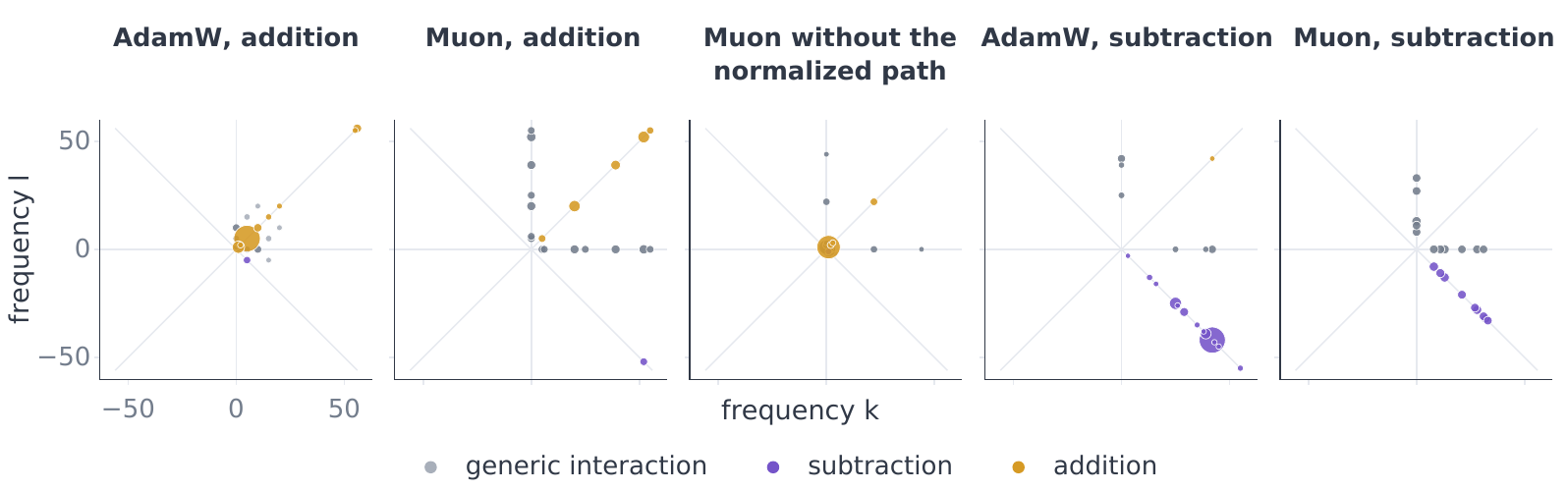}
\caption{The twenty highest-power conjugate pairs of the final residual at converged checkpoints, on the mode grid, sized by share of non-constant power; grey lines mark $l=k$ and $l=-k$. AdamW places $80.1\%$ of non-constant power on a single pair of the family the task selects; Muon's strongest pair carries $7.9\%$, with the rest spread across the family and beyond it; ablating the normalized path returns the strongest pair to $58.9\%$. Between the addition and subtraction panels the occupied line moves from one diagonal to the other.}
\label{fig:grid}
\end{figure}

\begin{table}[t]
\centering\small
\caption{Spectral dispersion at solved checkpoints of the matched depth-$1$ runs. Effective
pairs is $\exp(-\sum_i q_i\log q_i)$, with normalized pair powers $q_i$, over all $6{,}384$ non-constant conjugate pairs,
reported as a mean with range. Family power is the task-aligned family's share of
non-constant power, meaning $(k,k)$ for addition and $(k,-k)$ for subtraction, and top-five
power is the share carried by the five strongest pairs of the spectrum. The final row ablates the normalization and orthogonalization together.}
\label{tab:sparsity}
\begin{tabular}{llrrr}
\toprule
Operation & Optimizer & Effective pairs & Family power & Top-five power \\
\midrule
Addition & AdamW & $4.95$ \ \ $[3.68,\,10.20]$ & $91.0\%$ & $93.7\%$ \\
Addition & Muon & $326.09$ \ \ $[96.91,\,578.90]$ & $28.0\%$ & $13.1\%$ \\
\midrule
Subtraction & AdamW & $2.54$ \ \ $[2.45,\,2.72]$ & $95.4\%$ & $96.5\%$ \\
Subtraction & Muon & $211.46$ \ \ $[33.25,\,377.24]$ & $27.7\%$ & $16.0\%$ \\
\midrule
Addition & Muon, path ablated & $4.11$ \ \ $[2.04,\,6.05]$ & $66.5\%$ & $94.0\%$ \\
\bottomrule
\end{tabular}
\end{table}

Table~\ref{tab:sparsity} reports the entropy effective count $\exp(-\sum_i q_i\log q_i)$ over all $6{,}384$ non-constant conjugate pairs and the task-aligned family's power share on matched depth-1 runs. Here $q_i$ denotes normalized pair power. Across the five-seed study AdamW occupies $4.40$ effective pairs on average over thirty solved checkpoints, Muon $385.85$ over twenty-five, and Stable Muon $343.96$ over thirty-six: freezing the interface leaves the dispersion where Muon put it and removes the instability anyway. Within each family's $(p-1)/2$ pairs, the generality analysis instead uses the inverse participation ratio $1/\sum_i q_i^2$, with $q_i$ renormalized within the family: $1.99$--$2.59$ for solved AdamW variants and $21.33$--$55.34$ for Muon. On the uniform operand grid, a predictor using only the opposite family and a constant offset has accuracy exactly $1/p$, since sum and difference are independent for odd $p$. Across solved matched checkpoints, AdamW places $0.0$--$7.8\%$ of non-constant power there and Muon $2.5$--$17.7\%$.

\section{The matched branch through failure, recovery, and masking}
\label{app:branchtraj}

\begin{table}[t]
\centering\small
\caption{The addition family along matched branches. Pairs counts how many of the twenty
highest-power conjugate pairs at the final residual belong to the family; share is the
family's fraction of non-constant power; addition-only is accuracy when the representation
is filtered to the family.}
\label{tab:coordinates}
\begin{tabular}{lrrrrr}
\toprule
Step & Branch & Pairs & Share & Addition-only & Full \\
\midrule
$126{,}200$ & control & $11$ & $83.95\%$ & $100.00\%$ & $94.07\%$ \\
$126{,}200$ & freeze & $11$ & $83.95\%$ & $100.00\%$ & $94.07\%$ \\
\midrule
$160{,}000$ & control & $1$ & $11.88\%$ & $2.65\%$ & $3.29\%$ \\
$160{,}000$ & freeze & $12$ & $89.33\%$ & $100.00\%$ & $98.97\%$ \\
$180{,}000$ & control & $0$ & $0.32\%$ & $0.88\%$ & $0.97\%$ \\
$180{,}000$ & freeze & $12$ & $88.32\%$ & $100.00\%$ & $97.95\%$ \\
$230{,}000$ & control & $4$ & $67.69\%$ & $100.00\%$ & $98.41\%$ \\
$230{,}000$ & freeze & $12$ & $90.78\%$ & $100.00\%$ & $99.47\%$ \\
$300{,}000$ & control & $3$ & $20.38\%$ & $100.00\%$ & $45.85\%$ \\
$300{,}000$ & freeze & $12$ & $90.69\%$ & $100.00\%$ & $98.89\%$ \\
\bottomrule
\end{tabular}
\end{table}

Table~\ref{tab:coordinates} follows the family's top-twenty membership, power share, isolated accuracy, and full-model accuracy along the depth-4 matched pair of \S\ref{sec:outvoted} from a bit-identical state. Over $174{,}000$ steps, the frozen branch retains high filtered accuracy and substantial family power; the control loses all three at filtered-pathway failure through its current readout, recovers all three, and loses the share again at projection masking while the isolated family still gives $100\%$. The $\alpha$-sweep crossing at $\alpha=2.55$ occurs at a share of $0.6247$, below the frozen branch's native $0.9069$. Replaying a depth-1 collapse and recovery at ten-step resolution: full accuracy returns to $99.90\%$ within $300$ steps while the addition-family power cosine to the pre-collapse state reaches only $0.55$; the pre-collapse representation holds between $4.2\%$ and $5.7\%$ under the recovered readout throughout, while the recovered representation reads at $90.73\%$ under the pre-collapse readout. The matched branches' five strongest modes are identical at the branch point and share at most one pair at sampled steps $160{,}000$ through $230{,}000$. Patching the addition component of the frozen branch's final MLP write into the control under the frozen readout gives $99.99\%$; under the control's own readout, $5.05\%$; patching its full final MLP write gives $97.61\%$ and $4.46\%$ under the two readouts; patching only the non-addition component of that MLP write into the control gives $36.79\%$, below the control's unmodified $45.85\%$.

\section{Depth}
\label{app:depth}

\begin{table}[t]
\centering\small
\caption{Time to sustained generalization, in optimizer steps and in elapsed seconds on a
single device. Per-step cost is the median over evaluation intervals. The AdamW baseline at
depth $4$ is right-censored at the step budget.}
\label{tab:wallclock}
\begin{tabular}{llrrrr}
\toprule
Depth & Optimizer & ms/step & Steps to grok & Seconds to grok & Advantage \\
\midrule
\multirow{2}{*}{$2$} & AdamW & $33.06$ & $5{,}100$ & $170$ & \\
 & Muon & $57.81$ & $2{,}300$ & $133$ & $2.22\times$ steps, $1.28\times$ time \\
\midrule
\multirow{2}{*}{$4$} & AdamW & $61.68$ & $>300{,}000$ & $>19{,}768$ & \\
 & Muon & $110.87$ & $52{,}600$ & $5{,}856$ & $>5.70\times$ steps, $>3.37\times$ time \\
\bottomrule
\end{tabular}
\end{table}

\begin{table}[t]
\centering\small
\caption{Stage-level interventions at depth $4$, step $53{,}000$, baseline $99.30\%$.
Relocation transports the addition family onto other diagonal frequencies at matched power;
phase scrambling randomizes the phase of each conjugate pair. The two stochastic columns
show the reported realization (replicate $4$ of five); zeroing is deterministic.}
\label{tab:stagewise}
\begin{tabular}{llrrr}
\toprule
Stage & Kind & Relocation & Phase & Zeroed \\
\midrule
Block 1 attention write & write & $99.30\%$ & $99.30\%$ & $86.31\%$ \\
Block 2 attention write & write & $99.33\%$ & $99.29\%$ & $0.94\%$ \\
Block 3 attention write & write & $99.29\%$ & $99.30\%$ & $0.44\%$ \\
\midrule
Block 3 MLP activation & activation & $1.66\%$ & $0.00\%$ & $1.19\%$ \\
Block 3 MLP write & write & $0.95\%$ & $0.00\%$ & $1.19\%$ \\
Block 4 attention write & write & $3.58\%$ & $0.00\%$ & $58.44\%$ \\
\bottomrule
\end{tabular}
\end{table}

\begin{table}[t]
\centering\small
\caption{Direct-readout accuracy by post-block residual, decoding with the final unembedding and
skipping the remaining blocks. Values are ranges over the same $41$ checkpoints used in
Section~\ref{sec:depth}, each with native full-grid accuracy of at least $95\%$. Both columns
report percentages to two decimal places.}
\label{tab:depthreadout}
\begin{tabular}{llrr}
\toprule
Depth & Regime & Earlier blocks & Final block \\
\midrule
$2$ & AdamW & $0.72$--$0.88\%$ & $99.99$--$100.00\%$ \\
$2$ & Muon & $0.85$--$5.09\%$ & $95.83$--$99.95\%$ \\
$2$ & Stable Muon & $1.28$--$5.23\%$ & $96.37$--$99.40\%$ \\
$4$ & Muon & $0.78$--$17.97\%$ & $95.07$--$99.30\%$ \\
$4$ & Stable Muon & $0.69$--$17.58\%$ & $95.25$--$98.71\%$ \\
\bottomrule
\end{tabular}
\end{table}

\begin{figure}[!htbp]
\centering
\includegraphics[width=\textwidth]{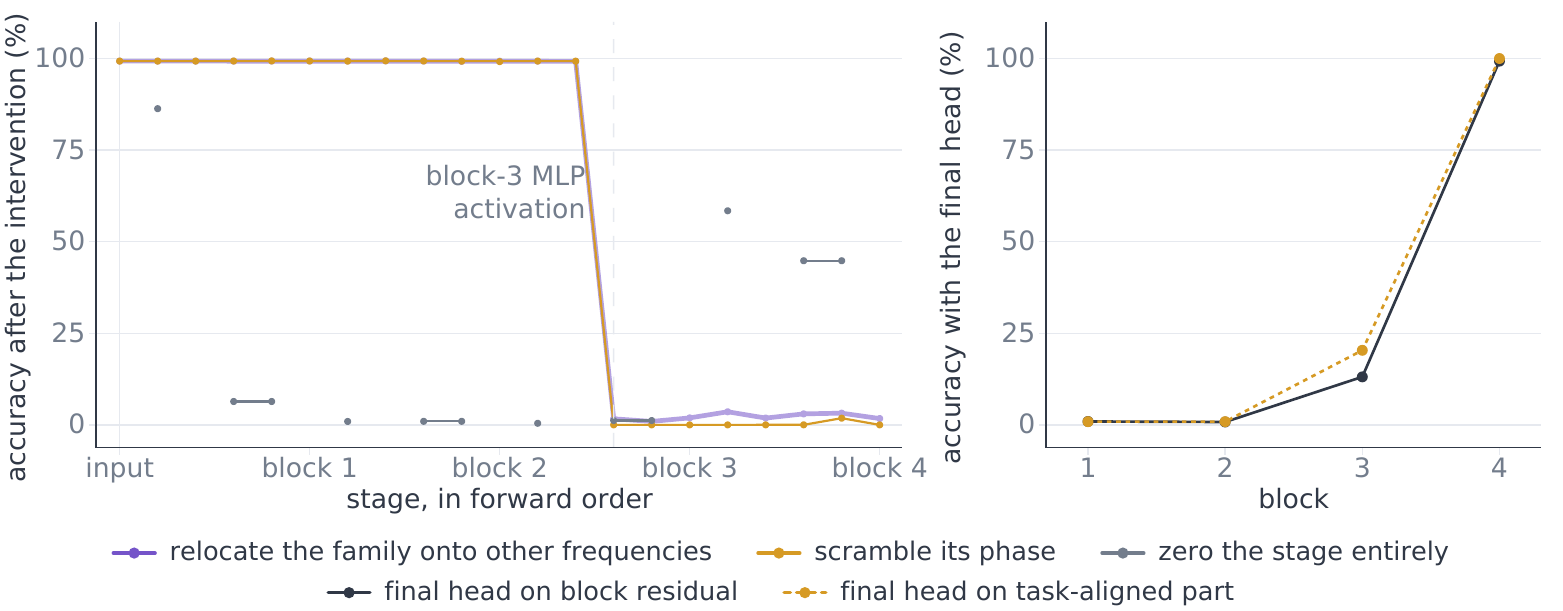}
\caption{Depth-$4$ Muon at step $53{,}000$. \textbf{Left:} the reported stage-intervention
realization. Relocation and phase scrambling first cause a large drop at the block-$3$ MLP
activation. The two stochastic curves use replicate $4$ of five and illustrate sensitivity;
individual later-stage outcomes vary across perturbations. Zeroing the earlier block-$2$ and
block-$3$ attention writes also destroys accuracy, despite their low sensitivity to these
Fourier perturbations. \textbf{Right:} direct decoding through the final unembedding, using
each post-block residual or its addition-filtered component. Both reach high accuracy only
at the final block.}
\label{fig:depth}
\end{figure}

Muon reaches sustained generalization in $2{,}300$ steps at depth 2 against $5{,}100$, and in $52{,}600$ at depth 4 where AdamW peaks at $96.69\%$ and never sustains the threshold within $300{,}000$ steps. Table~\ref{tab:wallclock} compares steps and elapsed time; a Muon step costs $1.75$--$1.80\times$ an AdamW step. At the depth-4 checkpoint in Table~\ref{tab:stagewise}, zeroing the block-2 and block-3 attention writes leaves $0.94\%$ and $0.44\%$ accuracy. Frequency relocation and phase scrambling at these writes preserve near-baseline accuracy. Both perturbations cause a large drop at the block-3 MLP activation. Figure~\ref{fig:depth} and Table~\ref{tab:stagewise} show the reported perturbation realization. Later-stage outcomes vary across the five perturbations, so near-chance accuracy in this realization does not establish uniform sensitivity across perturbations.

Across eleven solved checkpoints, the first stage where mean pair-phase scrambling costs more than half the baseline accuracy is an MLP activation in ten and an MLP write in one. This criterion averages five perturbations per stage. Decoding intermediate residuals with the model's final head measures compatibility with that particular head. Under this head, accuracy reaches $95\%$ only at the final block across all $41$ checkpoints selected by full-grid accuracy, never exceeding $17.97\%$ earlier (Table~\ref{tab:depthreadout}). The depth-$4$ Stable Muon checkpoint at step $210{,}000$ reaches $18.11\%$ at an earlier residual, with native full-grid accuracy $92.41\%$, placing it outside this solved-checkpoint subset. Within the eleven-checkpoint stage-intervention subset, among seventeen residual states with direct-readout accuracy between $2\%$ and $95\%$, family filtering improves sixteen, with a mean increase of $4.88$ percentage points across all seventeen. Across eighteen depth-4 checkpoints after the shared state, between-branch power cosine stays above $0.98$ in the first block; in the fourth it lies between $0.03$ and $0.44$ at fifteen checkpoints. The exact-state branch at step $126{,}200$ raises mean post-branch accuracy from $48.07\%$ to $97.50\%$, final accuracy from $36.88\%$ to $98.87\%$, and cuts evaluations below $90\%$ from $1{,}353$ to $66$.

\section{Decoder, basis, and arithmetic follow-ups}
\label{app:followup}

The decoder and matched-branch tests use the original architecture, data splits, and optimizer settings; operation and normalization controls are specified separately below. The basis study adds $52{,}698$ baseline updates and $9{,}000$ continuation updates. The subsequent causal study adds $364{,}302$ updates: five $24{,}000$-step arithmetic branches, fifteen $5{,}000$-step specificity branches, four original-trajectory extensions totaling $49{,}302$ steps, and four $30{,}000$-step operation or architecture controls. A subsequent horizon extension adds $420{,}000$ updates: each of the five corrected addition branches and the corrected subtraction run continues from step $30{,}000$ to $100{,}000$. The hidden-learning-rate control adds $470{,}000$ updates across five continuations from step $6{,}000$ to $100{,}000$. Verification replays are excluded from these totals. Multiple checkpoints and interventions from one trajectory are dependent observations.

\subsection{Monitoring and matched states}
\label{app:followup-protocol}

The causal branches and fresh baseline replications use a joint event: training and test accuracy must both fall below $90\%$ after sustained grokking. Training accuracy is checked at every update; test accuracy is checked every $100$ updates and at every post-grokking state with training accuracy below $90\%$. Thus every joint event is detected, while brief test-only excursions can fall between evaluations. Six consecutive grid evaluations at or above $95\%$ confirm grokking. The confirmation step is the \emph{last} evaluation of this six-check streak. Original sweep speeds instead report the start of a five-check streak (\S\ref{sec:setup}); the criteria therefore differ as well as their reporting points.

Within each seed, matched arithmetic branches restore identical parameters, optimizer buffers, and random-number-generator state. Exact next-update replays reproduce all three. The original prefixes combine historical evaluation grids with densely captured failures; their positive events establish binary incidence. We therefore compare event presence, without comparing their counts, durations, or historical minima against the uniformly monitored branches. The five trajectories were selected for mechanistic study and give no estimate of collapse prevalence over an unrestricted seed population.

The first captured joint failures occur at steps $17{,}494$, $17{,}721$, $18{,}069$, $16{,}308$, and $16{,}056$ for seeds $0$--$4$. Seeds $0$ and $4$ use exploratory dense replays from steps $16{,}000$ and $15{,}000$, respectively. Seed $0$'s earlier $100$-step log first detected failure at $17{,}500$; the grid missed seed $4$'s brief early failures. Seeds $1$--$3$ were specified together prospectively with the monitoring above and $200$ additional updates after capture. The selected failed decoder snapshots for seeds $3$ and $4$ are later states, $16{,}309$ and $16{,}060$, and retain those identities throughout.

\begin{table}[htbp]
\centering
\small
\caption{Test accuracy (\%) at adjacent captured failures. $h_0,W_0$ precede the update; $h_1,W_1$ follow it. Each head is evaluated with each feature state, without retraining. Original events are captured under mixed historical monitoring grids (Appendix~\ref{app:followup-protocol}).}
\label{tab:adjacent}
\begin{tabular}{lrrrr}
\toprule
Seed & Before: $W_0h_0$ & Old head: $W_0h_1$ & New head: $W_1h_0$ & After: $W_1h_1$ \\
\midrule
$0$ & $95.49$ & $96.39$ & $31.86$ & $31.87$ \\
$1$ & $94.14$ & $94.18$ & $13.56$ & $9.09$ \\
$2$ & $99.99$ & $100.00$ & $1.01$ & $0.93$ \\
$3$ & $89.54$ & $91.01$ & $50.80$ & $53.60$ \\
$4$ & $91.34$ & $98.66$ & $66.75$ & $66.33$ \\
\bottomrule
\end{tabular}
\end{table}

\subsection{Fresh decoders and the basis hypothesis}
\label{app:basis}

Fresh bias-free linear decoders are fitted to the raw final residual using only the original $3{,}830$ training examples. An uncentered SVD preconditioner retains all $128$ dimensions, with singular values floored at $10^{-6}$ of the largest. Each inner fit estimates its own preconditioner. A $20\%$ split of the training set selects an $L_2$ penalty on the preconditioned coefficients from $\{0,10^{-8},10^{-6},10^{-4},10^{-2}\}$ by validation cross-entropy; the selected decoder is then refitted on all training examples. The $8{,}939$ test examples enter only the final evaluation. Fits use float64 logistic regression and L-BFGS-B; convergence and unregularized controls are recorded.

At the selected collapsed checkpoints, decoder accuracies for seeds $0$--$4$ are $98.72\%$, $98.20\%$, $100.00\%$, $98.77\%$, and $100.00\%$, against $99.51\%$, $99.56\%$, $100.00\%$, $99.40\%$, and $100.00\%$ at their references. Probe checkpoints are steps $17{,}494$, $17{,}721$, $18{,}069$, $16{,}309$, and $16{,}060$: the last two occur after the adjacent transitions in Table~\ref{tab:adjacent}. These results establish substantial surviving linear task information; the small gaps in three seeds leave room for reduced decodability or probe limitations.

For column residuals, an exact global basis account requires one invertible $R$ satisfying $h_1(x)=Rh_0(x)$ across the operand grid. With examples stacked as rows, full column rank makes this equivalent to equality of the two feature column spaces. We test train-fitted linear and affine maps, centered reconstruction, rank sensitivity, principal angles, and a separately identified full-domain oracle. Centering by training means prevents a large constant feature component from dominating reconstruction error.

\begin{table}[htbp]
\centering
\small
\caption{Centered held-out reconstruction error (\%) for a train-fitted map from later to earlier residuals. Healthy controls match interval length. Reference intervals span $6$--$508$ updates and occur at different optimization stages, so comparisons are within each seed.}
\label{tab:followup-geometry}
\begin{tabular}{lrrr}
\toprule
Seed & \shortstack{Reference to\\captured failure} & Healthy interval & \shortstack{Adjacent\\failure update} \\
\midrule
$0$ & $43.63$ & $32.66$ & $1.88$ \\
$1$ & $43.05$ & $10.82$ & $4.63$ \\
$2$ & $9.05$ & $2.76$ & $1.10$ \\
$3$ & $45.73$ & $14.30$ & $10.66$ \\
$4$ & $25.60$ & $1.18$ & $25.63$ \\
\bottomrule
\end{tabular}
\end{table}

Positive controls plant exact residual gauges into the trained network. Eight controls cover condition numbers $1$, $3$, $10$, and $100$ in float32 and float64; twenty-four further controls plant observed fitted matrices with their measured orientation and conditioning. Exact compensation and train-fitted compensation recover $100\%$ in every fixed-condition control. For the three fresh adjacent transitions, observed forward-map errors are $4.65\%$, $1.10\%$, and $10.77\%$, against $0.0004$--$0.0006\%$ for their planted controls. These errors reject exact global equivalence for the measured pairs. Healthy evolution also departs from exact equivalence (Table~\ref{tab:followup-geometry}).

Approximate transport remains useful: inverse-linear compensation reaches $95.13\%$, $100\%$, and $94.37\%$ at the three fresh first events. Reference timing matters. Seed $0$'s head from step $17{,}200$ scores $14.05\%$ on the collapsed residual, while the head from $17{,}493$ scores $96.39\%$. Its geometric discrepancy from the distant reference is already present before the final failing update. Moreover, the complete real task-only Fourier coefficient matrices have shape $113\times128$; when both have full row rank, an invertible map between them always exists. For the seed-$4$ generalization-drop pair at steps $16{,}000$ and $17{,}000$, a fitted task-only map aligns that projection exactly while scoring $4.62\%$ on raw held-out features. This pair is separate from the adjacent failures above.

One-time repairs also have limited persistence. Eighteen continuations resume two captured failures for $500$ updates under nine head/state interventions each. Every successful repair retaining optimizer state falls below $90\%$ training accuracy on the next update. Clearing the head's Adam state improves eventual recovery but permits the immediate relapse. These tests separate instantaneous decodability from stability under continued optimization.

\subsection{Arithmetic interventions}
\label{app:arithmetic}

The accurate loss evaluates $\log(1+\sum_{j\ne y}\exp(z_j-z_y))$ with \texttt{log1p} whenever the correct-class logit is maximal; other examples use ordinary cross-entropy. Model storage, backward matrix arithmetic, and Muon's update remain float32. This avoids cancellation in the small loss and correct-class derivative, following the numerical concerns of \citet{prieto2025grokking}. All five step-$6{,}000$ sources already satisfy the six-check grokking criterion and already contain measurable gradient error. The intervention acts on an existing post-grokking trajectory.

\begin{table}[htbp]
\centering
\small
\caption{Matched arithmetic branches from step $6{,}000$ through $30{,}000$. Original columns show a captured joint failure. Accurate-CE branches have no joint failure; all finish at $100\%$ train and test accuracy. Minima include the starting state.}
\label{tab:followup-arithmetic}
\begin{tabular}{lrrr}
\toprule
Seed & Original event step & \shortstack{Original event\\test (\%)} & \shortstack{Accurate minimum\\test (\%)} \\
\midrule
$0$ & $17{,}494$ & $31.872$ & $99.754$ \\
$1$ & $17{,}721$ & $9.095$ & $99.843$ \\
$2$ & $18{,}069$ & $0.929$ & $98.859$ \\
$3$ & $16{,}308$ & $53.597$ & $99.978$ \\
$4$ & $16{,}056$ & $66.327$ & $99.843$ \\
\bottomrule
\end{tabular}
\end{table}

The specificity study branches each seed at step $15{,}000$ and runs to $20{,}000$ with three corrections. Accurate CE gives $0/5$ joint failures; target-derivative repair gives $0/5$; per-example zero-sum projection gives $0/5$. Every original trajectory has a captured failure in this window. All fifteen corrected branches finish at $100\%$ test accuracy; the minimum measured test accuracy across these branches is $99.978\%$.

For stock logit gradient $g$, target repair retains each wrong-class entry and sets $g_y=-\sum_{j\ne y}g_j$, leaving the reported forward loss unchanged. Projection replaces $g$ by $g-\mathbf{1}(\mathbf{1}^{\top}g)/p$. Both corrections compute their class sums in float64 and cast the result back to float32. Derivative errors are relative $L_2$ errors against the accurate float64 derivative evaluated on the identical float32 logits. The readout-gradient reference casts this logit derivative to float32 before backpropagating through the shared float32 network. At the fixed seed-$4$, step-$16{,}000$ audit, target repair reduces relative derivative error from $32.02\%$ to $4.19\times10^{-7}$, whereas projection leaves $31.88\%$. At step $20{,}000$, projected derivatives retain $52.94$--$86.07\%$ relative error despite a class-sum residual below $2.1\times10^{-8}$ relative to the reference norm. Thus restoring the class-sum identity can prevent the observed failures over this window even with substantial remaining derivative error. Numerical feature inflation and this projection control build on \citet{liu2026glitching}.

At step $15{,}000$, stock logit-gradient errors are $10.56$--$21.86\%$ and readout-gradient errors $18.95$--$43.42\%$. Passing a class-common gradient error through the same saved Adam state produces a class-centered update difference: its centered norm fraction is $45.85$--$73.37\%$, versus $(2.8$--$4.1)\times10^{-16}$ for scalar SGD, at float64 roundoff. These descriptive norm ratios are nonadditive. Accurate-CE branches also have smaller feature means: original mean norms at step $15{,}000$ are $2.22$--$3.83$ times their matched corrected values.

\textbf{Persistence through the original training horizon.} We resume each corrected step-$30{,}000$ state through step $100{,}000$, matching the original depth-$1$ horizon. All model parameters, optimizer buffers, and the PyTorch random state are restored. Twelve direct updates agree bitwise with both the monitored runner and a six-plus-six resumed continuation for all six sources. Each extension adds $70{,}000$ updates, with training accuracy recorded at every state and test accuracy every $100$ steps and whenever training accuracy is below $90\%$. The final and minimum-accuracy checkpoints reproduce their logged metrics exactly.

\begin{table}[htbp]
\centering\small
\caption{Corrected-loss runs through step $100{,}000$. Accuracies are percentages. Minima combine the earlier prefix with the extension, counting the shared step-$30{,}000$ state once. Addition is corrected from step $6{,}000$; subtraction uses corrected CE from initialization and its minima start at grokking confirmation.}
\label{tab:long-horizon}
\begin{tabular}{lrrrr}
\toprule
Run & Monitoring start & Min.\ test & Final test & Joint event \\
\midrule
Addition, seed 0 & $6{,}000$ & $99.754$ & $100.000$ & none \\
Addition, seed 1 & $6{,}000$ & $99.843$ & $100.000$ & none \\
Addition, seed 2 & $6{,}000$ & $98.859$ & $100.000$ & none \\
Addition, seed 3 & $6{,}000$ & $99.978$ & $100.000$ & none \\
Addition, seed 4 & $6{,}000$ & $99.843$ & $100.000$ & none \\
Subtraction, seed 0 & $10{,}500$ & $96.879$ & $99.955$ & none \\
\bottomrule
\end{tabular}
\end{table}

All six extensions complete without a joint event. Thus correction suppresses the observed failures through the original training budget. This addresses postponement beyond step $30{,}000$ within that budget; behavior after $100{,}000$ and brief test-only excursions between scheduled evaluations remain outside the measurement. The original stock trajectories already have captured failures by $30{,}000$, which establishes their positive binary incidence at the longer horizon. Their historical monitoring supports no comparison of event counts or durations with these dense branches.

\subsection{Mean amplification and finite-step overshoot}
\label{app:mean-update}

At the checkpoints immediately before the five original failures, feature means over the full operand grid have norms $10{,}374$--$12{,}203$. The main-text intervention in Eq.~\ref{eq:followup-mean} uses the post-update representation's training mean and adds each component to the rescued logits $W_{U,0}h_1$.
Applying the training-estimated mean unchanged to test examples, the mean term alone gives $32.11\%$, $9.42\%$, $0.93\%$, $53.34\%$, and $66.07\%$ accuracy, close to the full failing updates in Table~\ref{tab:followup-arithmetic}. The centered term alone retains $95.26\%$, $92.48\%$, $100\%$, $87.20\%$, and $98.46\%$. Relative reconstruction error is below $3\times10^{-16}$. This evaluation-only intervention establishes local sufficiency of the product of feature mean and readout displacement. Feature growth and optimizer history remain coupled before that step.

Replacing gradients only at the failing update with accurate gradients preserves all five failures. The state accumulated beforehand matters. For each actual complete parameter displacement $\Delta\theta$, the accurate training gradient satisfies $\nabla L^{\top}\Delta\theta<0$, including each group's contribution. Scaling the displacement by $0.001$ or $0.01$ lowers training loss in every seed; taking the full step raises it sharply. Both interpolation endpoints reproduce saved parameters exactly. The captured updates locally descend and overshoot at finite scale.

\subsection{Feature-mean timelines in the archived seed study}
\label{app:mean-timeline}

\begin{figure}[htbp]
\centering
\includegraphics[width=\textwidth]{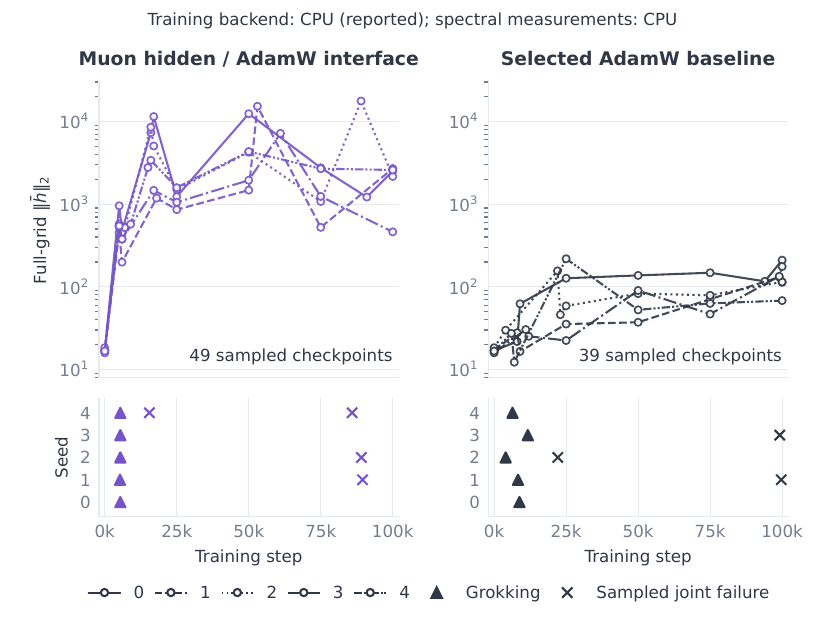}
\caption{Readout-input feature means in archived Muon and selected AdamW runs. Circles show all available checkpoints: $49$ for Muon and $39$ for AdamW. Lines connect these irregularly sampled states. Norms use the full operand grid. Lower rasters mark original grokking and joint failures observed on the $100$-update grid: four Muon failure observations across three seeds and three AdamW observations across three seeds. This sampling can miss brief events. The plot identifies configurations and the archive's reported CPU backend; no original runtime manifest is retained. These descriptive trajectories provide context for the matched mean-displacement and arithmetic interventions.}
\label{fig:mean-timeline}
\end{figure}

Figure~\ref{fig:mean-timeline} describes the available checkpoints from five original Muon and five selected AdamW trajectories. These archived trajectories are distinct from the fresh, densely captured cohort used for adjacent-update localization. For each checkpoint, we measure $\|\mu\|_2$, where $\mu=p^{-2}\sum_{a,b}h(a,b)$ is the mean of the actual readout-input features over the full operand grid. The archived orthonormal Fourier summaries give
\[
\|\mu\|_2=\frac{\sqrt{P_{\mathrm{total}} f_{\mathrm{DC}}}}{p},
\]
since $\widehat h(0,0)=p\mu$. This measurement uses the unfiltered representation. All $88$ plotted rows were independently recomputed; a direct feature-mean check on a saved checkpoint agrees with its orthonormal-DFT calculation to $5.49\times10^{-16}$ relative error.

Muon uses hidden learning rate $0.03$ and decay $0.1$, embedding AdamW learning rate $0.001$ and decay $1$, and readout AdamW learning rate $0.00025$ and decay $1$. The selected AdamW reference uses learning rate $0.001$ and decay $3$ throughout. The original archive reports CPU training, and its Fourier analysis runs on CPU. Checkpoint selection includes phase landmarks and low-accuracy states; intermediate feature means are unmeasured. The larger recorded Muon means are descriptive evidence about these configurations and sampled trajectories. The controlled intervention in \S\ref{sec:arithmetic} addresses the local effect of multiplying the observed readout displacement by the feature mean.

\subsection{Subtraction and normalized architecture}
\label{app:normalized}

Four additional runs use seed $0$ and a fixed $30{,}000$-step horizon: a stock/accurate-CE pair on subtraction and a pair on addition with gain-free RMS normalization before attention, before the MLP, and before the readout ($\epsilon=10^{-6}$). Parameter initialization and optimizer settings are matched, with no added parameters or hyperparameter search.

\begin{table}[htbp]
\centering
\small
\caption{Operation and architecture controls. Confirmation is the end of the six-check grokking streak. First-event test accuracies and post-grokking minima are separate measurements; both RMS runs subsequently recover to $100\%$. Each row is one seed.}
\label{tab:followup-generality}
\begin{tabular}{llrrrr}
\toprule
Condition & CE & Confirmation & First event & Event test (\%) & Min.\ test (\%) \\
\midrule
Subtraction & Stock & $9{,}300$ & $15{,}281$ & $5.851$ & $5.851$ \\
Subtraction & Accurate & $10{,}500$ & none & n/a & $96.879$ \\
RMS addition & Stock & $15{,}700$ & $18{,}515$ & $85.658$ & $0.481$ \\
RMS addition & Accurate & $19{,}900$ & $28{,}495$ & $0.884$ & $0.526$ \\
\bottomrule
\end{tabular}
\end{table}

Subtraction reproduces acute readout sensitivity: updated features with the preceding head score $98.81\%$, the mean term alone $5.84\%$, and the centered term $98.76\%$. At step $30{,}000$, accurate CE has no joint event and test accuracy is $99.966\%$. Its continuation through $100{,}000$ is reported in Table~\ref{tab:long-horizon}.

\begin{figure}[t]
\centering
\includegraphics[width=\textwidth]{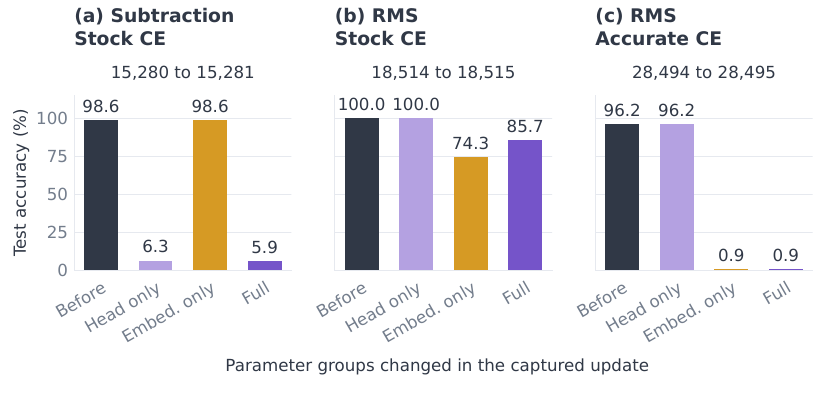}
\caption{One-step parameter interventions at the first detected joint failure in each operation or architecture control, all at seed $0$. Each bar uses the preceding checkpoint with only the indicated parameter group replaced by its following value; ``Full'' applies all three groups. In subtraction, the head update reproduces most of the damage. In both RMS cases, the embedding update is destructive while the head update preserves accuracy. The accurate-CE RMS failure occurs with an accurate loss derivative. These are first-event measurements at the displayed steps; the later post-grokking minima are reported separately in Table~\ref{tab:followup-generality}.}
\label{fig:followup-rms}
\end{figure}

In the RMS runs, the embedding update causes the acute loss of accuracy. In the accurate-CE event, test accuracy falls from $96.208\%$ to $0.884\%$; its logit derivative has relative error $9.48\times10^{-8}$ and no vanished correct-class entries. Updating only the head retains $96.208\%$, updating only embeddings gives $0.940\%$, and freezing the head still gives $0.895\%$. The minimum reaches $0.526\%$ at step $28{,}503$; the stock-RMS minimum is $0.481\%$ at $28{,}540$. Exact update replay passes. Both RMS event directions locally descend and lower loss at scales $0.001$ and $0.01$, then overshoot at full scale. This historical counterexample bounds the readout/feature-mean account to the measured unnormalized cases and rules out universal prevention by accurate loss arithmetic. Appendix~\ref{app:rms-replication} adds four prospective accurate-CE RMS seeds and applies the same event analysis to every captured failure.

\subsection{Post-grokking hidden-learning-rate reduction}
\label{app:lr-reduction}

We restore the same five original step-$6{,}000$ checkpoints used for the arithmetic branches, including the full model, optimizer buffers, and random-number-generator state. Immediately after restoration, we reduce the hidden Muon learning rate from $0.03$ to $0.003$ and hold it fixed through step $100{,}000$. Embeddings and readout remain trainable at learning rates $0.001$ and $0.00025$, respectively. All branches retain stock cross-entropy and the original weight-decay coefficients: $0.1$ for the hidden matrices and $1$ for embeddings and readout. Each continuation adds $94{,}000$ updates, for $470{,}000$ in total.

The primary outcome is whether a seed has any state with both training and held-out test accuracy below $90\%$. Training accuracy is recorded at every update, including the source and final states. Test accuracy is evaluated every $100$ updates and whenever training accuracy is below $90\%$, matching the corrected-loss protocol. This gives $94{,}001$ recorded training states per seed. Test-only excursions between scheduled evaluations can be missed.

\begin{table}[htbp]
\centering\small
\caption{Hidden-LR continuations through step $100{,}000$. Every seed has a joint failure. Test accuracies are percentages, and minima include the starting state. Below-$95\%$ counts include all test evaluations under the stated monitoring rule; the scheduled subset uses only the $100$-step grid. These counts describe the new branches.}
\label{tab:lr-reduction}
\begin{tabular}{lrrrrr}
\toprule
Seed & \shortstack{First joint\\failure step} & \shortstack{Minimum\\sampled test} & \shortstack{Evaluations\\below $95\%$} & \shortstack{Scheduled\\subset} & Final test \\
\midrule
$0$ & $18{,}767$ & $0.8614$ & $105$ & $13$ & $100.0000$ \\
$1$ & $18{,}476$ & $4.7209$ & $88$ & $20$ & $100.0000$ \\
$2$ & $19{,}439$ & $0.8950$ & $135$ & $14$ & $99.6644$ \\
$3$ & $17{,}041$ & $4.8999$ & $52$ & $22$ & $99.8881$ \\
$4$ & $17{,}578$ & $0.8166$ & $103$ & $13$ & $100.0000$ \\
\bottomrule
\end{tabular}
\end{table}

All five reduced-LR branches experience a joint failure, compared with $5/5$ original stock trajectories and $0/5$ corrected-loss branches through the same horizon. We compare binary seed-level incidence because the original trajectories have different historical monitoring. The reduced-LR branches first fail between steps $17{,}041$ and $19{,}439$; seeds $0$, $2$, and $4$ reach approximately chance-level test accuracy ($1/113\approx0.885\%$). All recover to at least $99.6644\%$ final test accuracy. Their recovery illustrates why final accuracy alone misses these failures.

\begin{table}[!htbp]
\centering\small
\caption{Sampled diagnostics for the hidden-LR branches. Each seed has $95$ measurements at $1{,}000$-step intervals, including steps $6{,}000$ and $100{,}000$. Feature means cover the full operand grid. CE error is the relative $L_2$ error of the stock float32 training-logit derivative against the accurate float64 derivative at the same logits, expressed as a percentage. Maxima are over the sampled diagnostic states.}
\label{tab:lr-reduction-diagnostics}
\begin{tabular}{lrrrrrr}
\toprule
& \multicolumn{3}{c}{Feature-mean norm} & \multicolumn{3}{c}{CE derivative error (\%)} \\
\cmidrule(lr){2-4}\cmidrule(lr){5-7}
Seed & Start & Final & Maximum & Start & Final & Maximum \\
\midrule
$0$ & $310.8$ & $292.6$ & $4{,}282.5$ & $1.440$ & $0.030$ & $38.558$ \\
$1$ & $459.5$ & $505.3$ & $8{,}783.8$ & $1.500$ & $0.180$ & $46.273$ \\
$2$ & $355.1$ & $1{,}096.5$ & $8{,}100.4$ & $0.967$ & $22.837$ & $36.585$ \\
$3$ & $411.9$ & $955.8$ & $8{,}878.5$ & $2.228$ & $5.408$ & $48.431$ \\
$4$ & $434.1$ & $445.1$ & $3{,}596.0$ & $1.748$ & $0.027$ & $63.645$ \\
\bottomrule
\end{tabular}
\end{table}

Every reduced-LR branch exhibits substantial feature-mean growth and loss-derivative error at sampled states (Table~\ref{tab:lr-reduction-diagnostics}). The accompanying artifacts retain training-set feature means, raw trajectories, checkpoint replays, and the detailed audits. We have not localized the parameter group responsible for these reduced-LR failures, so the evidence establishes joint failure without attributing it to the acute readout mechanism identified in the original branches. Ordinary learning-rate reduction also scales the decoupled weight-decay step: the hidden per-update decay factor changes from $1-0.03\times0.1=0.997$ to $1-0.003\times0.1=0.9997$. The tested tenfold reduction is therefore a stabilization control. It is insufficient to provide the stability achieved by corrected loss arithmetic for these five matched continuations through step $100{,}000$.

\subsection{Initialization and memorization decoder controls}
\label{app:decoder-controls}

We apply the same decoder to initialization and one memorization-phase state of each of the five unnormalized trajectories in Appendix~\ref{app:basis}. Checkpoint selection uses the recorded native trajectory: the first of five consecutive scheduled $100$-step training evaluations at or above $99.9\%$, requiring test accuracy below $95\%$ at that state and a step before original grokking. For seeds $0$--$4$, this selects steps $200$, $400$, $200$, $100$, and $200$; the corresponding fifth training checks occur at $600$, $800$, $600$, $500$, and $600$. The selection rule, checkpoint identities, and validation indices were fixed before decoder fitting. The initial registration and its pre-fitting amendment to the paper's sustained-memorization criterion are retained.

Replaying each saved initialization through step $1{,}000$ reproduces the original model, optimizer, and CPU RNG states exactly, as well as all ten scheduled evaluation rows. These replays recover the selected early checkpoints without substituting a new trajectory. Their original five-check grokking starts are $5{,}500$, $5{,}500$, $5{,}500$, $5{,}300$, and $5{,}400$, respectively; the historical speed sweep has its own run identities.

Features are captured by a pre-hook on the readout, with no centering or Fourier filtering. All four checkpoints within each seed use the same internal validation indices. The bias-free multinomial logistic regression starts from zero and fits in float64. Every candidate estimates its uncentered, full-dimensional SVD transform using only its fitting subset, with a relative singular-value floor of $10^{-6}$. Validation cross-entropy selects the penalty from $\{0,10^{-8},10^{-6},10^{-4},10^{-2}\}$; exact ties select the smaller penalty. The selected model is refitted from zero on all original training examples with a newly fitted transform. L-BFGS-B uses $500$ candidate iterations and $1{,}000$ refit iterations, \texttt{ftol} $=10^{-10}$, \texttt{gtol} $=10^{-7}$, at most three times the iteration budget in function evaluations, $20$ line-search steps, and history size $10$. These settings match the existing decoder procedure.

\begin{table}[t]
\centering\small
\setlength{\tabcolsep}{3.2pt}
\caption{Training-only decoder controls on the five trajectories used for the reference and failed-state probes. All accuracies are percentages. Memorization is the first of five consecutive scheduled training evaluations at $\geq99.9\%$, before grokking and with test accuracy below $95\%$. Refit reports convergence and iterations; Cand. gives converged candidate fits out of five. The ten reference and failure results are unchanged.}
\label{tab:decoder-phase-controls}
\begin{tabular}{clrrrrcc}
\toprule
Seed & State & Step & Native train & Native test & Decoder test & Refit & Cand. \\
\midrule
0 & Initialization & 0 & 0.76 & 0.84 & 0.12 & Yes (10) & 3/5 \\
0 & Memorization & 200 & 99.95 & 0.08 & 0.02 & Yes (36) & 5/5 \\
0 & Reference & 17{,}200 & 100.00 & 99.12 & 99.51 & Yes (19) & 5/5 \\
0 & Failed & 17{,}494 & 48.25 & 31.87 & 98.72 & Yes (18) & 5/5 \\
\midrule
1 & Initialization & 0 & 0.81 & 0.92 & 0.12 & Yes (10) & 2/5 \\
1 & Memorization & 400 & 99.97 & 0.22 & 0.25 & Yes (33) & 5/5 \\
1 & Reference & 17{,}300 & 100.00 & 99.19 & 99.56 & Yes (19) & 5/5 \\
1 & Failed & 17{,}721 & 15.67 & 9.09 & 98.20 & Yes (17) & 5/5 \\
\midrule
2 & Initialization & 0 & 0.73 & 0.89 & 0.07 & Yes (10) & 3/5 \\
2 & Memorization & 200 & 99.95 & 0.49 & 0.55 & Yes (34) & 5/5 \\
2 & Reference & 18{,}000 & 100.00 & 100.00 & 100.00 & Yes (19) & 5/5 \\
2 & Failed & 18{,}069 & 0.99 & 0.93 & 100.00 & Yes (18) & 5/5 \\
\midrule
3 & Initialization & 0 & 0.86 & 0.91 & 0.23 & Yes (11) & 3/5 \\
3 & Memorization & 100 & 100.00 & 0.09 & 0.07 & Yes (37) & 5/5 \\
3 & Reference & 15{,}800 & 100.00 & 99.07 & 99.40 & Yes (19) & 5/5 \\
3 & Failed & 16{,}309 & 42.64 & 32.12 & 98.77 & Yes (18) & 5/5 \\
\midrule
4 & Initialization & 0 & 1.04 & 0.83 & 0.04 & Yes (10) & 2/5 \\
4 & Memorization & 200 & 99.95 & 1.02 & 1.12 & Yes (33) & 5/5 \\
4 & Reference & 16{,}050 & 100.00 & 100.00 & 100.00 & Yes (19) & 5/5 \\
4 & Failed & 16{,}060 & 21.78 & 18.13 & 100.00 & Yes (17) & 5/5 \\
\bottomrule
\end{tabular}
\end{table}

The selected decoder reaches $0.04$--$0.23\%$ held-out accuracy at initialization and $0.02$--$1.12\%$ at memorization, compared with $99.40$--$100\%$ at the existing references and $98.20$--$100\%$ at the failed states (Table~\ref{tab:decoder-phase-controls}). All $20$ selected refits converge. Unregularized memorization refits also converge; all five initialization refits reach the $1{,}000$-iteration limit, with $0.18$--$0.62\%$ held-out accuracy. Candidate objectives, gradient norms, iteration counts, and termination messages accompany the saved coefficients and transforms. Independently recomputing training and test predictions from these arrays reproduces all $40$ new accuracy measurements.

The early controls support the interpretation that the failed checkpoints retain task information made accessible by training. They leave its precise emergence between checkpoints unresolved. Low early accuracy under this bounded decoder procedure does not establish mathematical absence of task information, and high failed-state accuracy alone does not identify a complete arithmetic algorithm or exact preservation of the earlier representation.

\subsection{Accurate-CE RMS replication and event localization}
\label{app:rms-replication}

\textbf{Registered cohort and execution.} The historical accurate-CE RMS run in Appendix~\ref{app:normalized} is a seed-$0$ CPU pilot with a $30{,}000$-step horizon. We extend it with prospectively registered seeds $1$--$4$ on CPU, preserving the pilot's backend and PyTorch $2.13.0$ implementation. All runs use addition modulo $113$, the original seeded $30\%$ split, one block, width $128$, four heads, MLP width $512$, ReLU, and bias-free projections. Gain-free RMS precedes attention, the MLP, and the readout, with $\epsilon=10^{-6}$. The model and all five Newton--Schulz iterations remain float32. Hidden Muon uses learning rate $0.03$, momentum $0.95$, Nesterov updates, and decay $0.1$. Embedding and readout AdamW use learning rates $0.001$ and $0.00025$, respectively, decay $1$, betas $(0.9,0.999)$, and source $\epsilon=10^{-8}$. Source values for the remaining optimizer options are retained. The cancellation-resistant cross-entropy is active from initialization.

The execution manifest records Python $3.14.6$, NumPy $2.5.2$, SciPy $1.18.0$, macOS ARM64 with Accelerate BLAS, one PyTorch intra-op and inter-op thread, deterministic algorithms, and highest float32 matrix-multiplication precision. Initial and event replays reproduce the historical seed-$0$ model exactly under this environment. Hashed source snapshots, the complete optimizer configuration, and the registered protocol accompany the study.

\textbf{Monitoring and capture.} Training accuracy is inspected at every state, using exact correct/total counts. Testing occurs every $100$ updates and at every post-confirmation state with training accuracy below $90\%$. Six consecutive scheduled test evaluations at or above $95\%$ confirm grokking on the sixth check. Each prospective run stops at its first subsequent joint train/test failure below $90\%$, or at step $100{,}000$. Outcomes distinguish confirmed grokking with failure, confirmed grokking with event-free completion, failure to confirm grokking, and technical interruption. Minima and evaluation counts over the post-confirmation window include the confirmation evaluation itself.

Checkpoints preserve initialization, confirmation, every $1{,}000$ updates, and the terminal state. After confirmation, the immediately preceding model, optimizer, and RNG states are retained for event capture. Prospective checkpoints include PyTorch CPU, Python, and NumPy RNG states; GPU RNG is unused and explicitly recorded as such. The historical checkpoint format preserves PyTorch RNG only. Complete-update replay verifies bitwise model and optimizer equality, the available RNG states, and the resulting event measurements. Any interruption requires exact saved-state recovery with documented replay overlap.

\textbf{Identical event analysis.} For every captured transition, we compare the accurate float32 loss derivative with an analytic float64 reference on the identical stored float32 training logits. Saved arrays support absolute $L_2$ and maximum error, relative $L_2$ error, reference norm, and class-sum residuals. All eight old/new combinations of embeddings, hidden matrices, and readout are evaluated on the original training and test splits, with exact accuracy counts and accurate CE. Fresh decoders are fitted immediately before and after the update, using the actual input to the unembedding after final RMS normalization. The fitting, validation, and convergence protocol is identical to Appendix~\ref{app:decoder-controls}, with validation indices shared within each seed and unregularized sensitivity fits retained.

\textbf{Historical pilot reanalysis.} At steps $28{,}494\to28{,}495$, held-out accuracy changes from $8{,}600/8{,}939=96.21\%$ to $79/8{,}939=0.88\%$. All four swaps retaining the preceding embeddings score $96.21$--$96.41\%$; all four using the updated embeddings score $0.88$--$0.94\%$. The pre-update derivative's relative $L_2$ error is $9.48\times10^{-8}$, with absolute error $4.06\times10^{-10}$ against reference norm $4.28\times10^{-3}$. Its class-sum residual has $L_2$ norm $4.30\times10^{-10}$. Exact complete-update replay passes.

The training-only decoder gives $97.37\%$ before this update and $1.58\%$ afterward; both validation-selected refits converge. The unregularized control gives $95.83\%$ before and $2.62\%$ afterward, with the latter reaching the $1{,}000$-iteration limit. Thus this pilot does not support the high failed-state linear accessibility found in the unnormalized cohort. The bounded fits leave open whether another linear fitting procedure could recover more information. The historical run's later recovery and minimum remain separately reported in Table~\ref{tab:followup-generality}.

\textbf{Prospective outcomes.} All four new seeds confirm grokking and reach a joint failure, at steps $15{,}095$, $15{,}789$, $14{,}271$, and $15{,}136$ for seeds $1$--$4$, respectively (Table~\ref{tab:rms-replication-outcomes}). These runs add $60{,}291$ training updates. No run finishes event-free, fails to confirm grokking, or has a technical interruption. Every terminal checkpoint reproduces its recorded metrics, and every complete-update event replay passes the model, optimizer, and RNG checks.

\begin{table}[htbp]
\centering\small
\setlength{\tabcolsep}{4pt}
\caption{Accurate-CE RMS outcomes. Seed $0$ is the historical CPU pilot; seeds $1$--$4$ form the prospective CPU cohort. Prospective endpoints are first joint failure or step $100{,}000$. The pilot continued to $30{,}000$; its displayed endpoint is the separately analyzed first event.}
\label{tab:rms-replication-outcomes}
\begin{tabular}{lrrl}
\toprule
Seed & Confirmation & Event / endpoint & Outcome \\
\midrule
$0$ (pilot) & $19{,}900$ & $28{,}495$ & Joint failure \\
$1$ & $11{,}500$ & $15{,}095$ & Joint failure \\
$2$ & $12{,}600$ & $15{,}789$ & Joint failure \\
$3$ & $10{,}000$ & $14{,}271$ & Joint failure \\
$4$ & $12{,}600$ & $15{,}136$ & Joint failure \\
\bottomrule
\end{tabular}
\end{table}

\begin{table}[htbp]
\centering\small
\setlength{\tabcolsep}{4pt}
\caption{Adjacent RMS events and training-only decoders. Accuracies are percentages; before means one update before the displayed event. Decoder inputs include final RMS normalization. The last column gives selected-refit convergence before/after. Seed $0$ is the historical pilot; seeds $1$--$4$ are prospective.}
\label{tab:rms-event-decoders}
\begin{tabular}{lrrrrrl}
\toprule
Seed & Event step & \multicolumn{2}{c}{Native test} & \multicolumn{2}{c}{Decoder test} & Converged \\ & & Before & After & Before & After & \\
\midrule
$0$ & $28{,}495$ & 96.21 & 0.88 & 97.37 & 1.58 & Yes / Yes \\
$1$ & $15{,}095$ & 99.70 & 1.63 & 100.00 & 17.35 & Yes / Yes \\
$2$ & $15{,}789$ & 100.00 & 54.55 & 100.00 & 64.86 & Yes / Yes \\
$3$ & $14{,}271$ & 100.00 & 62.86 & 100.00 & 93.82 & Yes / Yes \\
$4$ & $15{,}136$ & 97.87 & 1.03 & 98.15 & 2.65 & Yes / Yes \\
\bottomrule
\end{tabular}
\end{table}

Figure~\ref{fig:rms-replication} summarizes the complete parameter swaps and decoder results.

At each prospective event, replacing only embeddings is sufficient for joint failure. With old embeddings, every combination of hidden matrices and readout retains at least $97.87\%$ test accuracy. The hidden update partly offsets the embedding damage in seeds $2$ and $3$: embeddings-only accuracy is $53.73\%$ and $57.93\%$, while applying both embedding and hidden updates gives $54.55\%$ and $62.86\%$. Readout replacement changes test accuracy by at most $0.06$ percentage points across these paired combinations. The embedding update therefore supplies the acute damage at all four captured transitions, with smaller interactions from the other groups.

All selected decoder refits before and after the five analyzed events converge. In the prospective cohort, decoder accuracy falls from $98.15$--$100\%$ before the update to $2.65$--$93.82\%$ afterward. Seed $3$ retains substantial accessibility, whereas seeds $1$ and $4$ have much lower recovery. Thus the RMS events establish an architectural boundary for both the readout locus and the uniformly high failed-state decoder recovery in the unnormalized cohort. They leave the upstream cause of the embedding displacement unresolved. Tables~\ref{tab:rms-decoder-diagnostics} and~\ref{tab:rms-derivatives} retain convergence and derivative diagnostics; full solver records and arrays accompany the paper.

Each prospective run has one sampled post-confirmation test evaluation below $95\%$, its terminal joint failure. Including confirmation, the four runs have $37$, $33$, $44$, and $27$ post-confirmation test evaluations. Their minimum sampled test accuracies equal their terminal accuracies: $1.63\%$, $54.55\%$, $62.86\%$, and $1.03\%$. These event-stopped summaries have different horizons from the pilot's subsequent recovery and from the $100{,}000$-step unnormalized continuations.

\begin{table}[htbp]
\centering\small
\setlength{\tabcolsep}{4pt}
\caption{All eight parameter-group swaps at captured RMS failures. Seed $0$ is the historical pilot; seeds $1$--$4$ are prospective. Mask order is embeddings, hidden matrices, readout; $0$ retains the preceding state and $1$ uses the following state. Accuracy is in percent. CE is accurate float64 evaluation of the resulting float32 logits. Raw records additionally retain exact counts and float32 accurate CE.}
\label{tab:rms-swaps-1}
\begin{tabular}{lrrrrr}
\toprule
Seed & EHR & Train & Test & Train CE & Test CE \\
\midrule
$0$ & 000 & 95.87 & 96.21 & 0.34207 & 0.32198 \\
$0$ & 001 & 95.87 & 96.21 & 0.33814 & 0.32056 \\
$0$ & 010 & 96.50 & 96.41 & 0.13595 & 0.20983 \\
$0$ & 011 & 96.55 & 96.41 & 0.13393 & 0.2092 \\
$0$ & 100 & 1.15 & 0.94 & 6.7289 & 6.7765 \\
$0$ & 101 & 1.12 & 0.93 & 6.7313 & 6.7788 \\
$0$ & 110 & 1.20 & 0.89 & 6.808 & 6.8589 \\
$0$ & 111 & 1.17 & 0.88 & 6.811 & 6.8617 \\
\addlinespace
$1$ & 000 & 99.53 & 99.70 & 0.0091396 & 0.0054243 \\
$1$ & 001 & 99.53 & 99.71 & 0.008633 & 0.0051797 \\
$1$ & 010 & 100.00 & 100.00 & 2.9798e-05 & 3.6247e-05 \\
$1$ & 011 & 100.00 & 100.00 & 2.945e-05 & 3.5891e-05 \\
$1$ & 100 & 1.10 & 1.61 & 5.8687 & 5.8818 \\
$1$ & 101 & 1.10 & 1.61 & 5.8695 & 5.8827 \\
$1$ & 110 & 1.07 & 1.63 & 5.8567 & 5.8698 \\
$1$ & 111 & 1.07 & 1.63 & 5.8575 & 5.8707 \\
\addlinespace
$2$ & 000 & 100.00 & 100.00 & 6.4328e-05 & 5.5268e-05 \\
$2$ & 001 & 100.00 & 100.00 & 6.324e-05 & 5.477e-05 \\
$2$ & 010 & 100.00 & 100.00 & 2.8588e-05 & 3.2462e-05 \\
$2$ & 011 & 100.00 & 100.00 & 2.857e-05 & 3.2467e-05 \\
$2$ & 100 & 54.57 & 53.73 & 3.1082 & 3.105 \\
$2$ & 101 & 54.57 & 53.73 & 3.1082 & 3.105 \\
$2$ & 110 & 55.22 & 54.55 & 3.0872 & 3.0822 \\
$2$ & 111 & 55.22 & 54.55 & 3.0872 & 3.0823 \\
\addlinespace
\bottomrule
\end{tabular}
\end{table}

\begin{table}[htbp]
\centering\small
\setlength{\tabcolsep}{4pt}
\caption{All eight parameter-group swaps at captured RMS failures. Seed $0$ is the historical pilot; seeds $1$--$4$ are prospective. Mask order is embeddings, hidden matrices, readout; $0$ retains the preceding state and $1$ uses the following state. Accuracy is in percent. CE is accurate float64 evaluation of the resulting float32 logits. Raw records additionally retain exact counts and float32 accurate CE.}
\label{tab:rms-swaps-2}
\begin{tabular}{lrrrrr}
\toprule
Seed & EHR & Train & Test & Train CE & Test CE \\
\midrule
$3$ & 000 & 100.00 & 100.00 & 0.00091077 & 0.0010679 \\
$3$ & 001 & 100.00 & 100.00 & 0.00090878 & 0.0010657 \\
$3$ & 010 & 100.00 & 100.00 & 0.00026494 & 0.00030755 \\
$3$ & 011 & 100.00 & 100.00 & 0.00026434 & 0.00030686 \\
$3$ & 100 & 58.96 & 57.93 & 2.0143 & 2.1057 \\
$3$ & 101 & 58.96 & 57.93 & 2.014 & 2.1054 \\
$3$ & 110 & 64.15 & 62.86 & 1.8055 & 1.8808 \\
$3$ & 111 & 64.15 & 62.86 & 1.8052 & 1.8805 \\
\addlinespace
$4$ & 000 & 97.65 & 97.87 & 0.14149 & 0.12772 \\
$4$ & 001 & 97.65 & 97.87 & 0.13947 & 0.12727 \\
$4$ & 010 & 99.37 & 98.36 & 0.04257 & 0.083994 \\
$4$ & 011 & 99.40 & 98.41 & 0.041476 & 0.083846 \\
$4$ & 100 & 0.84 & 1.05 & 6.3598 & 6.3671 \\
$4$ & 101 & 0.84 & 1.05 & 6.3556 & 6.3626 \\
$4$ & 110 & 0.89 & 1.03 & 6.3563 & 6.3669 \\
$4$ & 111 & 0.89 & 1.03 & 6.3517 & 6.3621 \\
\addlinespace
\bottomrule
\end{tabular}
\end{table}

\begin{table}[htbp]
\centering\small
\setlength{\tabcolsep}{4pt}
\caption{Accurate-CE derivative diagnostics on identical stored float32 training logits. Seed $0$ is the historical pilot; seeds $1$--$4$ are prospective. Errors compare the full mean-loss derivative with the analytic float64 reference. Class-sum residual is the $L_2$ norm across examples of the sum over classes. Maximum absolute errors and reference class-sum residuals are retained in the supporting records.}
\label{tab:rms-derivatives}
\begin{tabular}{llrrrr}
\toprule
Seed & State & Relative error & Absolute error & Reference norm & Class-sum residual \\
\midrule
$0$ & Before & $9.48\times10^{-8}$ & $4.06\times10^{-10}$ & $4.28\times10^{-3}$ & $4.30\times10^{-10}$ \\
$0$ & After & $5.08\times10^{-8}$ & $8.83\times10^{-10}$ & $1.74\times10^{-2}$ & $1.55\times10^{-9}$ \\
$1$ & Before & $8.12\times10^{-8}$ & $1.00\times10^{-10}$ & $1.24\times10^{-3}$ & $1.18\times10^{-10}$ \\
$1$ & After & $3.88\times10^{-8}$ & $6.35\times10^{-10}$ & $1.63\times10^{-2}$ & $1.45\times10^{-9}$ \\
$2$ & Before & $4.76\times10^{-8}$ & $1.21\times10^{-12}$ & $2.54\times10^{-5}$ & $3.67\times10^{-13}$ \\
$2$ & After & $4.86\times10^{-8}$ & $6.02\times10^{-10}$ & $1.24\times10^{-2}$ & $9.58\times10^{-10}$ \\
$3$ & Before & $8.99\times10^{-8}$ & $1.92\times10^{-12}$ & $2.13\times10^{-5}$ & $9.97\times10^{-13}$ \\
$3$ & After & $5.04\times10^{-8}$ & $5.98\times10^{-10}$ & $1.19\times10^{-2}$ & $8.80\times10^{-10}$ \\
$4$ & Before & $5.28\times10^{-8}$ & $1.40\times10^{-10}$ & $2.64\times10^{-3}$ & $2.36\times10^{-10}$ \\
$4$ & After & $4.26\times10^{-8}$ & $7.16\times10^{-10}$ & $1.68\times10^{-2}$ & $1.37\times10^{-9}$ \\
\bottomrule
\end{tabular}
\end{table}

\begin{table}[htbp]
\centering\small
\setlength{\tabcolsep}{4pt}
\caption{RMS decoder convergence and unregularized sensitivity. Seed $0$ is the historical pilot; seeds $1$--$4$ are prospective. Candidate convergence reports successful validation-candidate fits out of five; selected-refit convergence is in Table~\ref{tab:rms-event-decoders}. Unregularized accuracy is held-out percent. Complete termination messages, objectives, iteration counts, and gradient norms accompany every fit.}
\label{tab:rms-decoder-diagnostics}
\begin{tabular}{llrrrl}
\toprule
Seed & State & Selected $\lambda$ & Candidates & Unreg.\ test & Unreg.\ converged \\
\midrule
$0$ & Before & $1.00\times10^{-4}$ & 5/5 & 95.83 & Yes \\
$0$ & After & $1.00\times10^{-2}$ & 3/5 & 2.62 & No \\
$1$ & Before & $1.00\times10^{-8}$ & 5/5 & 100.00 & Yes \\
$1$ & After & $1.00\times10^{-2}$ & 5/5 & 11.05 & Yes \\
$2$ & Before & $0$ & 5/5 & 100.00 & Yes \\
$2$ & After & $1.00\times10^{-2}$ & 5/5 & 55.35 & Yes \\
$3$ & Before & $0$ & 5/5 & 100.00 & Yes \\
$3$ & After & $1.00\times10^{-6}$ & 5/5 & 88.66 & Yes \\
$4$ & Before & $1.00\times10^{-4}$ & 5/5 & 96.21 & Yes \\
$4$ & After & $1.00\times10^{-2}$ & 3/5 & 4.13 & No \\
\bottomrule
\end{tabular}
\end{table}

\clearpage
\section{Task structure in Fourier filtering}
\label{app:fourier-control}

\subsection{Full-family projection is an orbit average}

Write a Fourier mode as $\chi_{k,l}(a,b)=\exp(2\pi i(ka+lb)/p)$. Averaging its translations gives
\begin{equation}
\frac{1}{p}\sum_{t=0}^{p-1}\chi_{k,l}(a+t,b-t)
=\chi_{k,l}(a,b)\frac{1}{p}\sum_{t=0}^{p-1}\exp\!\left(\frac{2\pi i(k-l)t}{p}\right)
=\mathbf{1}_{k=l}\chi_{k,l}(a,b).
\end{equation}
Linearity proves Eq.~\ref{eq:orbit-average} for each feature channel. The surviving modes include $(0,0)$. For subtraction, averaging $(a+t,b+t)$ instead retains $k+l=0$ modulo $p$. With a linear head, $W_U P h=P(W_U h)$; an affine head also commutes with this average because averaging preserves its constant bias. Excluding the constant mode gives the orbit average minus the global mean.

The full-family filter therefore supplies the equivalence relation defining the answer. It can pool logits from training and held-out examples with that answer. This is part of the intervention itself, even though no labels are passed to the FFT routine. Its accuracy measures an answer-conditioned average and requires separate evidence to identify the algorithm implemented by the unfiltered model.

\subsection{A lookup-table memorizer}

We use the repository's five seeded splits, each with $3{,}830$ training and $8{,}939$ held-out examples at $p=113$. Let $T$ denote the training set and $e_s$ the one-hot vector for class $s$. The predictor stores $z(a,b)=e_{a+b}$ on $T$ and outputs zero on every unseen pair. If $n_s$ is the number of training examples with sum $s$,
\begin{equation}
(P_{\mathrm{add}}z)(a,b)=\frac{n_{a+b}}{p}e_{a+b}.
\end{equation}
Every answer orbit is covered in each split, so filtering yields $100\%$ accuracy. For a uniform split of $m=3{,}830$ out of $N=p^2$ pairs, an orbit is uncovered with probability $q=\binom{N-p}{m}/\binom{N}{m}$. A union bound gives $\Pr(\text{some orbit uncovered})\leq pq<2.89\times10^{-16}$. Before filtering, all held-out logits are tied. We use the smallest class index as the deterministic tie rule, matching the implementation's argmax; uniform random tie breaking has expected accuracy $1/113\approx0.885\%$.

\begin{table}[htbp]
\centering\small
\caption{Addition memorizer on the original seeded splits. Correct counts use the $8{,}939$ held-out examples. The filtered logits have a unique correct maximum on every example.}
\label{tab:memorizer}
\begin{tabular}{lrrrr}
\toprule
Seed & Raw correct & Raw accuracy (\%) & Filtered correct & Filtered accuracy (\%) \\
\midrule
$0$ & $79$ & $0.884$ & $8{,}939$ & $100.000$ \\
$1$ & $68$ & $0.761$ & $8{,}939$ & $100.000$ \\
$2$ & $79$ & $0.884$ & $8{,}939$ & $100.000$ \\
$3$ & $81$ & $0.906$ & $8{,}939$ & $100.000$ \\
$4$ & $77$ & $0.861$ & $8{,}939$ & $100.000$ \\
\bottomrule
\end{tabular}
\end{table}

The same control on subtraction reaches $100\%$ after the corresponding filter in all five seeds. Omitting the constant mode also retains perfect accuracy in these controls. A seeded bijection of output labels preserves the perfect filtered result when evaluated against those relabeled classes, because the projection still pools within answer orbits. Averaging training donors alone is sufficient; averaging only held-out donors gives zero logits for the memorizer.

These controls establish that full-family filtered accuracy can be perfect without a learned generalizing arithmetic algorithm. The $95.14\%$ filtered accuracy at AdamW step $3{,}000$ therefore requires independent evidence to establish when an internal arithmetic algorithm was learned. Evidence for surviving information at the collapsed checkpoints comes from the raw-feature decoder tests and adjacent-state swaps in \S\ref{sec:branches}.

\begin{figure}[t]
\centering
\includegraphics[width=\textwidth]{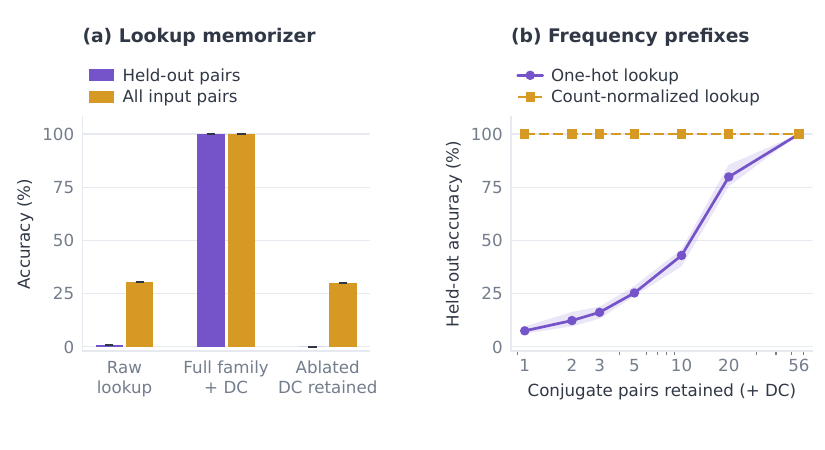}
\caption{Task structure supplied by Fourier filtering at $p=113$. \textbf{(a)} The lookup memorizes the $3{,}830$ training answers and returns zero on the $8{,}939$ held-out inputs. Full-family filtering with DC gives $100\%$ accuracy. Family ablation retains DC and yields $0\%$ held-out accuracy and $29.99\%$ full-grid accuracy. Bars show means over five splits; whiskers span their range. \textbf{(b)} Training-count normalization allows any one diagonal conjugate pair plus DC to classify every input correctly. The ordinary lookup's pair powers are mathematically tied; its prefixes use increasing frequency as a fixed tie break. Lines show means and shading spans the seed range.}
\label{fig:fourier-memorizer}
\end{figure}

\subsection{Sparse sufficiency, ablation, and perturbation controls}

A second lookup stores $(p/n_c)e_c$ for a training example with label $c$, using training-label counts only, and zero for unseen pairs. Its full projection is exactly $e_{a+b}$. Retaining any one nonzero diagonal conjugate pair and DC gives class-$c$ logits
\begin{equation}
z_k(s,c)=\frac{1+2\cos(2\pi k(s-c)/p)}{p},\qquad s=(a+b)\bmod p.
\end{equation}
For prime $p=113$ and nonzero $k$, the unique maximum is at $c=s$. All $56$ individual pairs give $100\%$ training, held-out, and full-grid accuracy on all five splits ($280$ checks against the closed form). Thus sparse filtered sufficiency also admits a lookup-only predictor. The construction can use a $113$-dimensional lookup representation with an identity head, padded to the transformer's $128$-dimensional residual width.

The ordinary one-hot lookup has equal mathematical power in all diagonal conjugate pairs. With a fixed ascending-frequency tie break, five-pair filtering gives $23.82$--$28.57\%$ held-out accuracy; sorting the numerically computed powers instead gives $63.69$--$87.46\%$. These ranges expose the sensitivity of the selected frequencies to floating-point tie breaking in this control. Figure~\ref{fig:fourier-memorizer} uses the fixed ordering. The count-normalized lookup reaches $100\%$ at every tested prefix size $1,2,3,5,10,20,56$.

The original family-ablation convention is $z-Pz+\bar z$, retaining DC. For both lookups, it preserves all training predictions and gives $0\%$ held-out accuracy, hence $3{,}830/12{,}769=29.9945\%$ full-grid accuracy. The trained models' near-chance full-grid ablations in Table~\ref{tab:swap} contain additional information absent from this simple control. Full-family sufficiency and low held-out ablation accuracy can nevertheless occur together under memorization.

We also apply $20$ fixed-seed replicates per split and lookup for pair-global phases, channelwise phases, and derangements of the $56$ pair coefficients, preserving DC, conjugate symmetry, and total nonconstant family power. For the ordinary lookup, mean held-out accuracies of the isolated family are $1.0755\%$, $0.9426\%$, and $3.0680\%$, respectively; for the count-normalized lookup they are $0\%$, $0.8850\%$, and $3.1555\%$. Each mean pools $20$ perturbations per split across five splits; raw replicate distributions are released. The sensitivity observed in the trained models remains a functional measurement of their projected components. These controls show why that sensitivity alone leaves the internal algorithm unidentified.

\subsection{Restricting the averaging donors at captured failures}

Using both endpoints of the five adjacent failure pairs in Table~\ref{tab:adjacent}, we recompute logits from the saved raw residuals and native heads, multiplying in float64. We average only held-out inputs with the same answer, excluding the evaluated input itself. This removes both training donors and self contribution. Input order, labels, and masks are checked against the original seeded splits.

\begin{table}[htbp]
\centering\small
\caption{Held-out accuracy (\%) before and after each captured update. ``Average'' uses held-out donors with the same answer and excludes the evaluated example. The operation still supplies the grouping.}
\label{tab:fourier-donors}
\begin{tabular}{lrrrr}
\toprule
 & \multicolumn{2}{c}{Before} & \multicolumn{2}{c}{After} \\
Seed & Raw & Average & Raw & Average \\
\midrule
$0$ & $95.49$ & $100.00$ & $31.87$ & $30.46$ \\
$1$ & $94.14$ & $100.00$ & $9.09$ & $0.93$ \\
$2$ & $99.99$ & $100.00$ & $0.93$ & $0.88$ \\
$3$ & $89.54$ & $99.12$ & $53.60$ & $65.29$ \\
$4$ & $91.34$ & $98.18$ & $66.33$ & $71.25$ \\
\bottomrule
\end{tabular}
\end{table}

Before these updates, high accuracy survives removal of training donors, establishing task-aligned signal in the logits of unseen examples beyond the ordinary lookup control. The donor-restricted operation still uses the answer orbits and remains a task-informed diagnostic. These checkpoints are distinct from the early AdamW checkpoint at step $3{,}000$; this analysis makes no claim about its donor dependence.

Independent FFT, translated-grid sums, and orbit reductions agree in float64 at moduli $7$ and $113$. We also check projector idempotence, constant-mode conventions, readout commutation, coverage of every answer orbit, and the single-pair closed form. All scripts, protocols, integer counts, perturbation outcomes, and input hashes accompany the release.

\section{Projection interference and spectral analyses}
\label{app:projection-detail}

\begin{figure}[!htbp]
\centering
\includegraphics[width=0.95\textwidth]{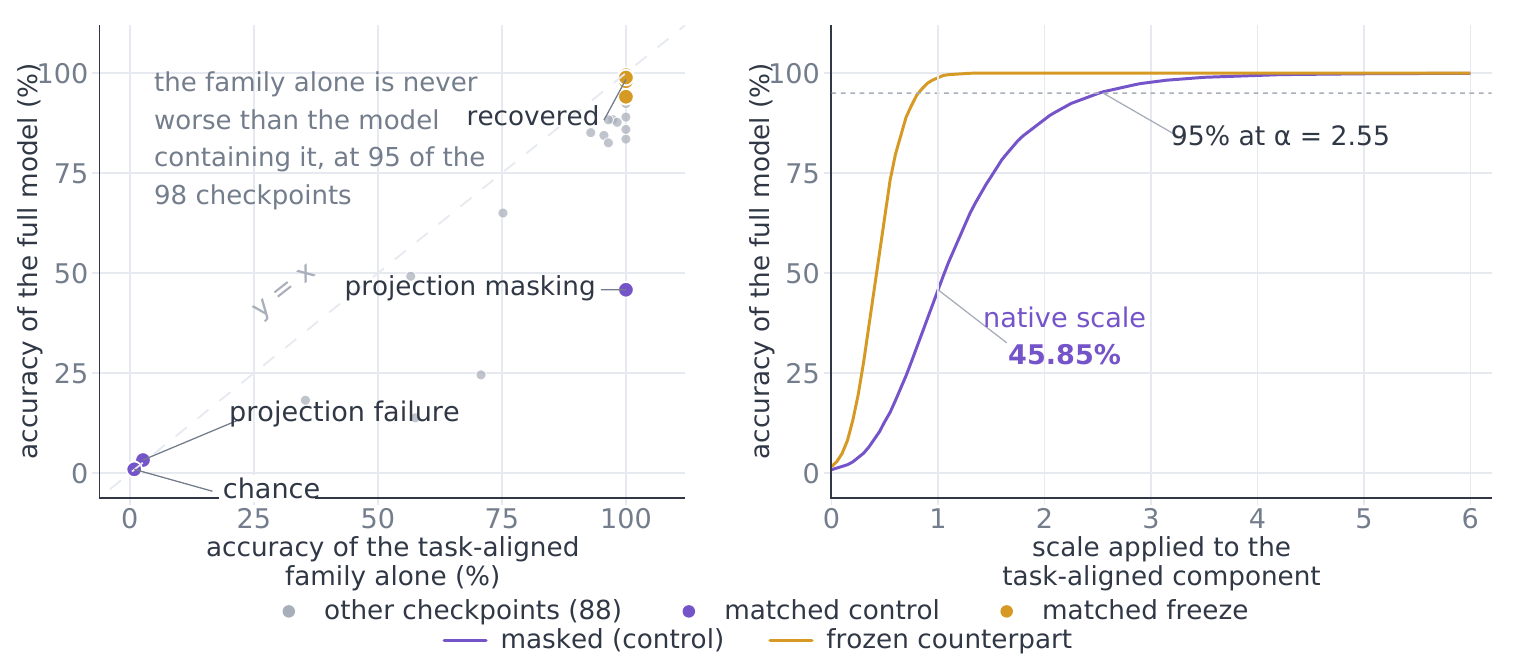}
\caption{\textbf{Left:} task-aligned family accuracy (including the retained constant mode) against full-model accuracy over $98$ checkpoints from the original studies, both evaluated through the current readout. \emph{Projection failure} denotes low filtered accuracy; \emph{projection masking} sits at the lower right, where the task-informed projection is accurate and the full model fails. At $95$ of $98$ checkpoints the family alone is at least as accurate as the full model, and the three exceptions exceed it by under a point. \textbf{Right:} rescaling the non-constant task-aligned component of the masked checkpoint by $\alpha$, with the rest of the representation and the readout untouched.}
\label{fig:modes}
\end{figure}

Filtering the final residual gives an answer-conditioned average through the current readout (Eq.~\ref{eq:orbit-average}). Comparing it with the full model measures interference from the remainder. In \emph{projection failure}, filtering gives $2.65\%$ and then chance, and the family has fallen out of the top twenty power pairs almost entirely. Fresh-decoder recovery in \S\ref{sec:branches} provides a separate test of surviving task information. In \emph{projection masking}, the filtered representation achieves $100\%$ accuracy over the full operand grid. The full model reaches $45.85\%$. The projection averages to a correct answer for every orbit; the full prediction also includes the remainder. One matched trajectory passes through failure, full recovery, and masking; its frozen counterpart retains high filtered accuracy and substantial family power (Appendix~\ref{app:branchtraj}).

\begin{table}[!htbp]
\centering
\caption{Correct-class margin at the masked checkpoint, decomposed by linearity of the bias-free readout. The competitor is the highest-scoring incorrect class under the full model, fixed across components. The constant mode belongs to the remainder. The final column is the fraction of examples with a positive component margin.}
\label{tab:margins}
\begin{tabular}{llrrrr}
\toprule
Branch & Component & Mean & On correct & On wrong & Fraction $>0$ \\
\midrule
Control & full & $-0.71$ & $2.15$ & $-3.13$ & $0.4585$ \\
Control & task-aligned & $+5.72$ & $5.30$ & $6.08$ & $1.0000$ \\
Control & remainder & $-6.43$ & $-3.15$ & $-9.21$ & $0.0475$ \\
\midrule
Freeze & full & $+11.61$ & $11.76$ & $-1.67$ & $0.9889$ \\
Freeze & task-aligned & $+19.28$ & $19.30$ & $17.18$ & $1.0000$ \\
Freeze & remainder & $-7.67$ & $-7.54$ & $-18.86$ & $0.0622$ \\
\bottomrule
\end{tabular}
\end{table}

\textbf{Masking is a competition on amplitude.} Because the unembedding is linear and bias-free, the correct-class margin decomposes exactly into the task-aligned contribution and the remainder (Table~\ref{tab:margins}). Against the full model's highest-scoring incorrect class, the non-constant family contributes a positive margin on every example in both branches, including every example the masked model answers wrongly, and on those examples its margin is higher, $+6.08$ against $+5.30$. The failures coincide with stronger opposition from the remainder: the remainder contributes negatively on more than nine tenths of examples in both branches, and in the control an opposition of $-6.43$ outvotes a task-aligned term of $+5.72$. Rescaling the task-aligned component by $\alpha$, with nothing retrained, raises the masked model monotonically on the sampled grid from $45.85\%$ to at least $95\%$ at $\alpha=2.55$ and $99.91\%$ at $\alpha=6$: the intervention changes the balance of the two logit contributions without retraining. The rescaled component is $P_{\mathrm{add}}h-\bar h$, with $\bar h$ the global mean. Recovery therefore depends on the task structure supplied by the projection. Family power share is a descriptive measure of this change.

\textbf{Family power during grokking.} Along an AdamW trajectory sampled every thousand steps, the addition family reaches $95.14\%$ full-grid filtered accuracy at step $3{,}000$, when the model's held-out test accuracy is $0.30\%$ and the family holds $1.05\%$ of non-constant power. Ablating the family leaves near-chance accuracy at every sampled checkpoint. Its share rises to $70.62\%$ at the test-accuracy jump, alongside greater concentration within the family. The memorizer control shows why this filtered accuracy alone cannot establish algorithm learning at step $3{,}000$. Increasing power share describes the trajectory and is compatible with the cleanup account of \citet{nanda2023progress}. The margin decomposition and rescaling intervention establish amplitude competition between projected logit components at the masked checkpoint.

\textbf{Spectral stability and functional change.} Exact nonzero Fourier support is invariant under an invertible residual-basis map; power and its rankings are invariant under orthogonal maps. General invertible maps can change dominant-frequency membership. A jointly compensated basis change preserves predictions. At the collapse in Figure~\ref{fig:collapse}, dominant-frequency Jaccard is $1.0000$ and power cosine is $0.9899$, while the filtered pathway's mean margin changes from $+18.14$ to $-46.85$. Over this interval the unembedding moves by $1.285$ of its norm and the addition-filtered residual by $0.392$; the old head on new features scores $26.31\%$, against $5.74\%$ for the reverse. Adjacent swaps provide finer localization (Figure~\ref{fig:followup}a). Accuracy returns to $99.90\%$ within $300$ steps with a changed power allocation. In the depth-4 pair, patching the frozen branch's addition component of the final MLP write into the control gives $99.99\%$ with the frozen head and $5.05\%$ with the control head. These results show readout sensitivity; exact basis equivalence requires the separate tests in Appendix~\ref{app:basis}.

\section{Computational cost}
\label{app:computational-cost}

We count arithmetic operations and stored scalars in the released dense implementations, at fixed precision. Let $p$ be the modulus and number of output classes, $T=3$ the sequence length, $d$ the residual width, $f\geq d$ the MLP width, $H$ the number of attention heads, $L$ the number of blocks, and $N$ the number of training examples. The dataset contains $p^2$ operand pairs. The bias-free model has
\[
P=L(4d^2+2df)+(2p+T+1)d
\]
parameters, including token and position embeddings and the readout.

\textbf{Model evaluation and training.}
Define
\[
\mathcal F(B)=B\bigl[L\{T(d^2+df)+T^2d\}+dp\bigr].
\]
A forward pass on $B$ examples costs $O(\mathcal F(B))$: the linear projections contribute the $Td^2$ and $Tdf$ terms, attention products contribute $T^2d$, and the final-token classifier contributes $dp$. Backpropagation has the same order. For a full-batch gradient update, model parameters, gradients, optimizer state, resident data, and saved activations occupy
\[
O\!\left(P+p^2T+NL\{T(d+f)+HT^2\}+Np\right).
\]
The explicit attention scores account for the $HT^2$ term. An unbatched evaluation without gradients on $B$ examples uses
$O(P+p^2T+B\{T(d+f)+HT^2+p\})$ storage; the training and evaluation peak is the larger of the two bounds. Gain-free RMS preserves these asymptotic orders.

\textbf{Optimizer updates.}
For one Muon matrix of shape $m\times n$, the implementation transposes when necessary so that $X$ has $r=\min(m,n)$ rows and $s=\max(m,n)$ columns. Each Newton--Schulz iteration computes $XX^\top$, a square of that $r\times r$ matrix, and a product with $X$. With $k$ iterations, the matrix update costs
\[
O\!\left(mn+k(r^2s+r^3)\right)
\]
and uses $O(mn+r^2)$ storage, including momentum and temporary matrices. Iterations run without autograd, so their number does not multiply the storage requirement. Summing over hidden matrices gives $O(P+kL(d^3+d^2f))$ work per optimizer step. AdamW takes $O(P)$ work and state for an all-parameter update; its auxiliary groups in the Muon configuration remain within this linear term. A run of $U$ updates therefore costs
\[
O\!\left(U\{\mathcal F(N)+P+kL(d^3+d^2f)\}
+\sum_e\mathcal F(B_e)\right),
\]
where the sum counts additional evaluations on $B_e$ examples. This expression excludes checkpoint I/O and the separate analyses below. With $T=3$ and $f=4d$, the training term reduces to $O(NLd^2+Ndp)$ per update and the Newton--Schulz term to $O(kLd^3)$.

\textbf{Fourier analysis.}
For a cached $p\times p\times c$ feature array, the two-dimensional FFT and each masked inverse FFT take $O(cp^2\log p)$ work under the standard FFT cost model and $O(cp^2)$ storage. Here $c=d$ for residuals and $c=f$ for MLP activations. Feature extraction additionally requires model evaluation on the grid. Applying the linear head to one filtered final-residual grid costs $O(p^3d)$ and materializes $O(p^3)$ logits in the unbatched implementation. Repeated filters incur these costs separately; interventions at earlier sites also require the remaining network computation. The depth extractor retains multiple sites, with up to $O(Lp^2T(d+f))$ cached feature storage.

\textbf{Training-only decoders.}
For one full-dimensional SVD-preconditioned fit with $n\geq d$ fitting examples, constructing and applying the preconditioner costs $O(nd^2+d^3)$. A multinomial loss-and-gradient evaluation costs $O(ndp)$. With $E$ such evaluations, including the initial objective call, fixed L-BFGS memory $q=10$, and the recorded bounded solver procedure, one fit costs $O(nd^2+d^3+Endp)$ and uses $O(nd+np+d^2+qdp)$ working storage. The five penalty candidates and final refit repeat this computation on their respective fitting rows; any additional sensitivity fits add the same costs. Retaining the full-grid features adds $O(p^2d)$ storage, and evaluating the fitted head adds the corresponding matrix-product and logit-storage costs. Solver effort is described by its realized evaluations and recorded convergence status. These are computational bounds for the implemented procedures; the study does not establish statistical sample-complexity or convergence-rate guarantees.

\section{Asset provenance and licensing}
\label{app:licensing}

The supplementary source archive includes \nolinkurl{LICENSING.md} and a \nolinkurl{licenses/} directory containing license texts, attribution notices, component versions, source locations, and SHA-256 records. The inventory covers the external Python imports in the released code, their active dependencies in the inspected environments, the Plotly JavaScript in the HTML artifacts, the bundled LaTeX files, and the fonts and tools used to produce the paper. Package versions observed during this audit are distinguished from versions retained in experiment protocols.

Third-party components retain their upstream terms and attribution. These include the BSD-family and MIT-family licenses used by the scientific Python packages, the Matplotlib license, the LaTeX Project Public License for the identified bibliography and header components, and the applicable font licenses. Distribution-specific bundled notices are retained alongside the package-level records. The matching documented source for the bundled \texttt{natbib.sty} is supplied to satisfy its source-distribution condition.

The task examples are generated from modular-arithmetic grids; the released measurements, checkpoints, and fitted decoders are produced by the reported experiments. The audited release has no project-wide reuse-license declaration for the original research assets. The inventory records that status explicitly. The conference templates are identified by their official source and the notices present in the supplied files. This documentation preserves the existing terms and grants no additional rights over either the original or third-party material.

\end{document}